\documentclass[10pt,twocolumn]{IEEEtran}

\usepackage{amsmath,amscd,amssymb,amsgen,amsfonts,amsbsy}
\usepackage{mathrsfs}
\usepackage[utf8x]{inputenc}
\usepackage{color}
\usepackage{textcomp}
\usepackage{float}
\usepackage{latexsym,graphicx}
\usepackage{tikz}
\usetikzlibrary{automata,arrows,positioning,calc}
\usepackage{url}
\usepackage{blkarray}
\usepackage{algorithmic}
\usepackage[ruled,lined]{algorithm2e}

\newcommand{\prob}[1]{\mathsf{Pr}\left( #1 \right)}

\newcommand{\remove}[1]{}

\newcommand{\comments}[1]{}

\newcommand{\expect}[1]{\mathsf{E}\left({#1}\right)}

\newtheorem{lemma}{Lemma}

\newtheorem{theorem}{Theorem}
\newtheorem{remark}{Remark}
\newtheorem{definition}{Definition}

\makeatother 

\title{Indexability of Finite State Restless Multi-Armed Bandit and Rollout Policy}

\author{
	Vishesh Mittal, Rahul Meshram,    Deepak Dev and  Surya Prakash \\ 
	IIIT Allahabad \\ 
	India
}

\begin{document}

\maketitle

\begin{abstract}
	We consider finite state restless multi-armed bandit problem. The decision maker can act on $M$ bandits out of $N$ bandits in each time step. The play of arm (active arm) yields state dependent rewards based on action and when the arm is not played, it also provides rewards based on the state and action. The objective of the decision maker is to maximize the infinite horizon discounted reward. 
	The classical approach to restless bandits is Whittle index policy. In such policy, the $M$ arms with highest indices are played at each time step. 
	Here, one decouples the restless bandits problem by analyzing relaxed constrained restless bandits problem. Then by Lagrangian relaxation problem, one decouples restless bandits problem into $N$ single-armed restless bandit problems.  We analyze the single-armed restless bandit. In order to study the Whittle index policy, we show structural results on the single armed bandit model. We define indexability and show indexability in special cases. We propose an alternative approach to verify the indexable criteria for a single armed bandit model using value iteration algorithm. We demonstrate the performance of our algorithm with different examples. We provide insight on condition of indexability of restless bandits using different structural assumptions on transition probability and reward matrices. 
	
	We also study online rollout policy and discuss the computation complexity of algorithm and compare that with complexity of index computation. Numerical examples illustrate that index policy and rollout policy performs better than myopic policy. 
\end{abstract}

\section{Introduction}
\label{sec:intro} 
Restless multi-armed bandit (RMAB) is a class of sequential decision problem. RMABs are extensively studied for resource allocation problems like---machine maintenance \cite{Glazebrook05,Glazebrook2006}, congestion control \cite{Avrachenkov13}, healthcare \cite{Mate21}, opportunistic channel scheduling\cite{Meshram18}, recommendation systems\cite{Meshram15,Meshram17}, queueing systems, \cite{Ansell03,Gittins11} etc.  RMAB problem is described as follows. There is a decision maker with $N$ independent arms where each arm can be in one of many finite states and the state evolves according to Markovian law. The playing of arm yields a state dependent reward and  we assume that even not playing of arm also yields state dependent reward. The transition dynamics of state evolution is assumed to be known  at the decision maker, The system is time slotted. The decision maker plays $M$ arms out of $N$ arms in each slot. The goal is to determine which sequence arms to be played such that it maximizes long-term discounted cumulative reward.


RMAB is the generalization of classical multi-armed bandit (Markov bandit). It is shown to be PSPACE hard \cite{Papadimitriou1999}.  RMAB is first introduced in  \cite{Whittle88} and author proposed Whittle index policy approach. This is a heuristic policy. The popularity of Whittle index policy is due to  optimal performance under some conditions  \cite{Weber90}. In such policy, the arms  with highest Whittle indices are played at each time-step. The index is derived for each arm by analyzing a single-armed restless bandit, where subsidy is introduced for not playing of arm and this is Lagrangian relaxation of restless bandit problem.  To use the Whittle index policy, one requires to show  indexability condition. Sufficient condition for indexability is provided under modeling assumption on restless bandits, particularly transition probability and reward matrices.  
In \cite{Bertsimas-Nino-Mora96,Bertsimas-Nino-Mora00}, authors introduced linear programming (LP) approach to restless bandits and proposed primal-dual heuristic algorithm. In \cite{Nino-Mora07}, author studied the marginal productivity index (MPI) for RMAB which is generalization of Whittle index policy and futher developed adaptive greedy policy to compute MPI. Their model assumes to satisfy partial conservation laws (PCL). In RMAB with PCL, one can apply adaptive greedy algorithm for Whittle index computation. 
In general, it is difficult to show the indexability and then compute the index formula. It is possible to compute the index approximately and this is done in \cite{Akbarzadeh2022}. The index computation algorithm using two timescale stochastic approximation approach and value iteration algorithm is studied in  \cite{Avrachenkov13,Borkar17b,Kaza19}. 
All this work requires to prove indexability and then compute the indices. Results on indexability make structural assumptions on transition probability and reward matrices.

A generalization of RMAB is referred to as \textit{weakly coupled Markov decision processes} (MDPs), where more complex constraints are introduced \cite{Hawkins03}, and approximate dynamic program is studied in \cite{Adelman08}. Fluid appproximation for restless bandits is studied in \cite{Bertsimas16}. Simulation-based approach for restless bandits is considered in \cite{Meshram20}.  Linear programming approach  with asymptotic optimality using fluid model is studied in \cite{Zhang2022}. Other paper on fluid model for resource allocation is \cite{Brown2022}.
Another work on LP policy for weakly coupled MDPs is studied in \cite{Gast2022}.
In \cite{NinoMora2020},  index computation algorithm is discussed. 
Recently, study of restless bandits is considered for partially observable states and hidden Markov bandit model in \cite{Kaza19,Meshram18,Meshram22}.

\subsection{Our contributions}
In this paper, we study finite state restless multi-armed bandit problem. We analyze the single-armed restless bandit problem. Our goal here is to show that the restless bandit is indexable. 
First, we  provide structural results by making modeling assumption on  transition probability and reward matrices and prove the indexability analytically. 

Second contribution in this paper is to study single-armed restless bandit using the value-iteration algorithm with fixed subsidy (Lagrangian parameter) and obtain the optimal decision rule for each state. In this, the optimal decision can be either passive action or active action. 
Next, we  vary subsidy over a fixed-size grid of subsidies and we compute the optimal decision rule for each state and for each value of subsidy from the fixed-size grid and it is referred to as the \textit{policy matrix}. By analyzing this matrix, we provide an alternative approach to verify the indexability condition. It also provides an index value for each state. The advantages of our approach is that is easy to verify the indexability condition using value iteration scheme. 
We present the analysis of indexability using policy matrix with few examples. 
Best of our knowledge, this viewpoint is not taken into study in earlier works. 

Third, we discuss  different numerical examples from \cite{Avrachenkov2022,Fu2019,Akbarzadeh2022,Nino-Mora07} and present examples of non-indexable bandits. We give insight on conditions for the non-indexability of restless bandits using numerical examples. We observe that when the reward is monotone (non-decreasing/non-increasing) for both actions in same direction, then bandit is indexable. When the reward for passive action is increasing in state, the reward for active action is decreasing in state and there is sufficient difference between the rewards of passive and active action and some structure on transition probability matrices, then the bandit is shown to be non-indexable. This example is motivated from \cite{Nino-Mora07}. However, for most of applications in communication systems opportunistic scheduling problem \cite{Borkar17b} and recommendation systems \cite{Meshram15,Meshram17}, the rewards are monotone for both actions in the same direction, thus the bandit is indexable.

Fourth, we compare the performance of Whittle index policy, myopic policy and  online rollout policy. Numerical examples illustrate that the discounted cumulative reward under index policy and rollout policy is higher than  that of myopic policy for non-identical restless bandits.    

The rest of the paper is organized as follows. RMAB problem is formulated in Section~\ref{sec:Model}. Indexability and our approach of indexability  for a single armed restless bandit is discussed in Section~\ref{sec:indexability}. 
Different algorithms for RMAB are presented in Section~\ref{sec:algo-rmab}. 
Numerical examples for indexability  of single-armed restless bandit are given in Section \ref{sec:numerical-indexability-sab}. Numerical examples for RMAB are illustrated in Section \ref{sec;numerical-example-rmab}. We make concluding remarks in Section~\ref{sec:concluding-remark}.

\section{Model of Restless Bandits}
\label{sec:Model}
Consider $N$ independent $K$ state Markov decision processes (MDPs). Each MDP has   state space  $\mathcal{S} =\{1,2, \cdots, K\}$ and action space $\mathcal{A} = \{0,1\}.$ Time progresses in discrete time-steps  and it is denoted by $t =0,1,2,3,\dots.$ 
Let $X_t^{n}$ denotes the state of Markov chain $n$ at instant $t \geq 0$ for $ 1 \leq n \leq N,$ and $X_t^{n} \in \mathcal{S}.$ Let $p_{i,j}^{a,n}$ be the transition probability from state $i$ to $j$ under action $a$ for $n$th Markov chain. Thus the transition probability matrix of $n$th MDP is $P^{a,n}= [[p_{i,j}^{a,n}]]$ for $a \in \mathcal{A}.$ Rewards from state $i$ under action $a$ for  $n$th  MDP is $r^{n}(i,a).$ 
Let $A_t^n$ be the action correspond to $n$th MDP at time $t,$ and $A_t^n \in \{0,1\}.$
Let $\pi_t: H_t \rightarrow \mathcal{M}$ is the strategy that maps  history to subset $\mathcal{M} \subset \mathcal{N},$ where $\mathcal{N} = \{1,2, \dots, N\}$ and $\vert\mathcal{M}\vert = M.$ Under policy 
$\pi,$ the action at time $t$ is denoted by 
\begin{equation*}
A_t^n = 
\begin{cases}
1 & \mbox{$n \in \mathcal{M},$}  \\ 
0 & \mbox{$n \notin \mathcal{M}.$}
\end{cases}
\end{equation*}
The infinite horizon discounted cumulative reward starting from initial state $X_0 = s$ with fixed policy $\pi$ is given as follows. 
\begin{equation*}
V^{\pi}(s) =  \mathrm{E}^{\pi}\left[\sum_{t=0}^{\infty}\beta^{t}\left(\sum_{n=1}^{N}r^{n}(X_t^n, A_t^n)\right)\right].
\end{equation*}
Here, $\beta$ is the discount parameter, $0<\beta<1.$
The agent can play at most $M$ arms at time $t.$ Thus,
\begin{equation*}
\sum_{n=1}^{N}  A_t^n = M, \ \  \forall \ \  t.
\end{equation*}
Our objective is 
\begin{eqnarray*}
	& \max_{\pi \in \Pi} V^{\pi}(s) \nonumber \\
	& \ \ \ \  \text{ subject to }  \sum_{n=1}^{N}  A_t^n = M, \forall t.
\end{eqnarray*}
It is the discounted (reward-based) restless bandit problem.
Here, $\Pi$ is a set of policies.
The relaxed discounted restless bandit problem is as follows. 
\begin{eqnarray}
& \max_{\pi \in \Pi} V^{\pi}(s) \nonumber \\
& \ \ \ \  \text{ subject to } \sum_{t=0}^{\infty} \beta^{t}\left(\sum_{n=1}^{N}  A_t^n \right) = \frac{M}{1-\beta}.
\label{eqn:relaxed-RMAB}
\end{eqnarray}
Then the Lagrangian relaxation of problem~\eqref{eqn:relaxed-RMAB} is given by 
\begin{eqnarray*}
	\max_{\pi \in \Pi} \mathrm{E}^{\pi}\left[\sum_{t=0}^{\infty}\beta^{t}\left(\sum_{n=1}^{N}r^{n}(X_t^n, A_t^n) + \lambda (1-A_t^n) \right)\right].   
\end{eqnarray*} 
Here, $\lambda$ is Lagrangian variable and $\lambda \in \mathbb{R}.$
This is equivalent to solving $N$ single-armed restless bandits, i.e., a MDP with finite number of states and two actions model and policy $\pi:\mathcal{S} \rightarrow \{0,1\}.$  Thus, a single-armed restless bandit problem is given as follows.
\begin{eqnarray}
\max_{\pi \in \Pi} \mathrm{E}^{\pi}\left[\sum_{t=0}^{\infty}\beta^{t}\left(r^{n}(X_t^n, A_t^n) + \lambda (1-A_t^n) \right)\right]
\end{eqnarray}
\subsection{Single-armed restless bandit}
For notation simplicity, we omit  superscript $n$ and introduce explicit dependence of value function on $\lambda.$ Then, the optimization problem from starting state $s$ under policy $\pi$ is as follows. 
\begin{eqnarray}
V(s, \lambda)   = \max_{\pi \in \Pi} \mathrm{E}^{\pi}\left[\sum_{t=0}^{\infty}\beta^{t}\left(r(X_t, A_t) + \lambda (1-A_t) \right)\right]
\end{eqnarray}
The optimal dynamic program   is given by 
\begin{eqnarray*}
	V(s, \lambda) = \max_{a \in \{0,1\}} \left\{ r(s,a) + (1-a) \lambda + \beta  \sum_{s^{\prime} \in \mathcal{S}} p_{s,s^{\prime}}^a V(s^{\prime},\lambda)  \right\}
\end{eqnarray*}
for all $s \in \mathcal{S}.$
We use these ideas to define the indexability of single-armed restless bandit in next section. 
The analysis  is done in the next section.

\section{Indexability and Whittle index}
\label{sec:indexability}
The Lagrangian variable $\lambda$ is referred to as subsidy. Then, we can rewrite dynamic program as follows.
\begin{eqnarray*}
	Q(s,a=0, \lambda) &=& r(s,a =0) + \lambda + \beta \sum_{s^{\prime} \in S} p_{s,s^{\prime}}^0 V(s^{\prime},\lambda) \nonumber \\
	Q(s,a=1, \lambda) &=& r(s,a =1)  + \beta \sum_{s^{\prime} \in S} p_{s,s^{\prime}}^1 V(s^{\prime},\lambda) \nonumber \\
	V(s, \lambda) &=& \max \left\{ Q(s,a=0, \lambda) , Q(s,a=1, \lambda)\right\} 
	\label{eqn:dynamic-prog-single-arm}
\end{eqnarray*}
for all $s \in \mathcal{S}.$
We define the passive set $B(\lambda)$ as follows.
\begin{eqnarray*}
	B(\lambda) = \bigg\{s \in \mathcal{S}: Q(s,a=0, \lambda)  \geq Q(s,a=1, \lambda) \bigg\}
\end{eqnarray*}
Let $\pi_{\lambda}: \mathcal{S} \rightarrow \mathcal{A}$  be the policy that maps state $s$ to action $a$ for given subsidy $\lambda.$ We can also rewrite  $B(\lambda) $ as 
\begin{eqnarray*}
	B(\lambda) = \bigg\{s \in \mathcal{S}: \pi_{\lambda}(s) = 0 \bigg\}.
\end{eqnarray*}

\begin{definition}[Indexable]
	A single armed restless bandit $(\mathcal{S}, \mathcal{A} = \{0,1\}, P^0, P^1, R, \beta)$ is \emph{indexable} if $B(\lambda)$ is non-decreasing in $\lambda$, i.e. 
	\begin{eqnarray*}
		\lambda_2 \geq \lambda_1 \implies B(\lambda_1) \subseteq  B(\lambda_2).  
	\end{eqnarray*}
	\end {definition}
	A restless multi-armed bandit problem is \emph{indexable} if each  bandit is \emph{indexable}.  A restless bandit is indexable if as the passive subsidy increases, then  the set of states for which the passive action is optimal increases.
	
	\begin{definition}[Whittle index]
		If a restless bandit $(\mathcal{S}, \mathcal{A} = \{0,1\}, P^0, P^1, R, \beta)$ is indexable  then its \emph{Whittle index}, $\lambda:\mathcal{S} \rightarrow R,$ is given by $\lambda(s) = \inf \{\lambda:s \in   B(\lambda )\},$ for $s \in \mathcal{S}.$
	\end{definition}
	We assume that the rewards  are bounded which implies  that the indices are also  bounded. 
	
	\begin{remark}
		\begin{enumerate} 
			\item  In general, it is very difficult to verify that  a bandit is indexable and even more  difficult to compute the Whittle indices. Note that as $\lambda$ increases, the set $B(\lambda)$ has to increase for indexability. 
			\item If a bandit is indexable, then it implies that there are $\lambda_1, \lambda_2 \in \mathbb{R},$  $\lambda_2 > \lambda_1$ such that $\pi_{\lambda_1}(s) = 1$ and $\pi_{\lambda_2}(s) = 0$ for $s \in \mathcal{S}$ because $B(\lambda_1) \subseteq  B(\lambda_2).$ This further suggests that the policy $\pi_{\lambda}(s)$ has a single threshold as a function of $\lambda$ for fixed $s \in \mathcal{S}.$ That is, there exist $\lambda^*$ such that 
			\begin{eqnarray*}
				\pi_{\lambda}(s) = 
				\begin{cases} 
					1 & \mbox{ $\lambda < \lambda^*,$ } \\
					0 & \mbox{ $\lambda \geq \lambda^*,$}
				\end{cases}
			\end{eqnarray*}
			for each $ s \in \mathcal{S}.$ 
			\item If a bandit is non-indexable, then it implies that there are  $\lambda_1, \lambda_2 \in \mathbb{R},$  $\lambda_2 > \lambda_1$ and $B(\lambda_1)$ is not a subset of  $ B(\lambda_2)$  for some  $\lambda_1, \lambda_2 \in \mathbb{R}.$ In other words, there exists some $\hat{s} \in \mathcal{S}$ such that for  $\lambda_3>\lambda_2>\lambda_1,$  $\pi_{\lambda_1}(\hat{s}) = 1,$ $\pi_{\lambda_2}(\hat{s}) = 0,$ and $\pi_{\lambda_3}(\hat{s}) = 1.$
		\end{enumerate}
		\label{remark-1}
	\end{remark} 
	
	\subsection{Single threshold policy and indexability}
	
	We now describe a single threshold type policy for  MDP having action space $\mathcal{A} = \{0,1\}$ in $s$  assuming  a fixed subsidy $\lambda$ as follows. We consider the deterministic Markov policy. The policy is called a single threshold type when the decision rule $\pi_{\lambda}(s)$ for $s \in \mathcal{S}$ is of the following form. 
	\begin{eqnarray*}
		\pi_{\lambda}(s) = 
		\begin{cases} 
			0 & \mbox{ $s \leq s^*,$} \\
			1 & \mbox{ $s > s^*,$}
		\end{cases}
	\end{eqnarray*}
	and $s^*$ is a threshold state. This is also referred to as \textit{control limit policy} in \cite{Puterman2005}. The conditions required on transition probability  and reward matrices for a single threshold policy are discussed below. 
	
	\begin{lemma}[Convexity of value functions]
		\begin{enumerate} 
			\item For fixed $\lambda$ and $\beta,$ $ V(s,\lambda),$ $  Q(s,a=0,\lambda),$  and $Q(s,a=1, \lambda)$ are non-decreasing piecewise convex in $s.$ 
			\item For fixed $s$ and $ \beta,$ $ V(s,\lambda),$ $ Q(s, a= 0,\lambda),$ and $Q(s, a=1,\lambda)$ are non-decreasing piecewise convex in $\lambda.$ 
		\end{enumerate}
		\label{lemma:convex}
	\end{lemma}
	Proof of lemma follows from induction method and it is given in Appendix~\ref{app:lemma:convex}. 
	
	\begin{definition} 
		The $Q(s,a,\lambda)$  is superadditive for a fixed $\lambda,$  if $s^{\prime}> s;$ $s^{\prime},s \in \mathcal{S}$ 
		\begin{equation}
		Q(s^{\prime},1, \lambda) +Q(s,0,\lambda) \geq Q(s,1,\lambda) +  Q(s^{\prime},0,\lambda) .
		\end{equation}
	\end{definition}
	This implies that $Q(s,1,\lambda) - Q(s,0, \lambda)$ is non-decreasing in $s.$  
	This is also referred to as \textit{Monotone non-decreasing difference property} in $s.$ This implies that the decision rule $\pi_{\lambda}(s)$ is non-decreasing in $s.$
	We now state the following result from \cite{Puterman2005}. 
	
	\begin{theorem}
		Suppose that
		\begin{enumerate}
			\item $r(s,a)$ is non-decreasing in $s$ for all $a \in \{0,1\},$
			
			\item $q(k\vert s,a) = \sum_{j=k}^{K} p_{s,j}^{a} $ is non-decreasing in $s$ for all $k \in \mathcal{S}$ and $a \in \{0,1\},$
			
			\item $r(s,a)$ is a superadditive (subadditive) function on $\mathcal{S} \times \{0,1\},$
			\item $q(k|s,a)$ is a superadditive (subadditive) function on $\mathcal{S} \times \{0,1\}$ for all $k \in \mathcal{S}.$
		\end{enumerate}
		Then, there exists optimal policy $\pi_{\lambda}(s)$ which is non-decreasing (non-increasing) in $s.$
		\label{thm:monotone-s}
	\end{theorem} 
	These assumptions imply that one requires stronger conditions on transition probability matrices $P^0,$ $P^1$ and reward matrix $R.$ 
	\begin{remark}
		\begin{enumerate}
			\item If $r(s,a)$ is a superadditive function on $\mathcal{S} \times \{0,1\},$ then  
			\begin{eqnarray*}
				r(s^{\prime},1) -r(s^{\prime},0)  \geq r(s,1) -r(s,0)
			\end{eqnarray*}
			for $s^{\prime} > s$ and $s^{\prime},s \in \mathcal{S}.$
			\item  If $q(k|s,a)$ is a superadditive function on $\mathcal{S} \times \{0,1\},$ then 
			\begin{eqnarray*}
				q(k|s^{\prime},1) - q(k|s^{\prime},0)  \geq  q(k|s,1)  -  q(k|s,0)
			\end{eqnarray*}
			for all $k\in \mathcal{S}.$
			This implies  
			\begin{eqnarray*}
				\sum_{j=k}^{K} \left[ p_{s^{\prime},j}^{1} - p_{s^{\prime},j}^{0 } \right] \geq 
				\sum_{j=k}^{K} \left[ p_{s,j}^{1} - p_{s,j}^{0 } \right].
			\end{eqnarray*}
		\end{enumerate}   
	\end{remark}
	
	From Theorem~\ref{thm:monotone-s}, it is clear that optimal policy is of a single threshold type in $s$ for a fixed $\lambda.$

	We next  prove that $Q(s,a,\lambda)$  is subadditive on $\{0,1\} \times \mathbb{R}$ for all values of $s \in \mathcal{S}.$ That is,  $Q(s,a=1,\lambda)- Q(s,a=0,\lambda)$ is non-increasing in $\lambda$ for each $s\in \mathcal{S}.$  Observe that  $Q(s,a=0,\lambda)$ is strictly increasing in $\lambda$ and  $Q(s,a=1,\lambda)$ is non-decreasing in $\lambda$  for fixed $s.$ For  assumptions in Theorem~\ref{thm:monotone-s}, $Q(s,a=1,\lambda)- Q(s,a=0,\lambda)$ is non-increasing in $\lambda.$ 
	Hence the policy $\pi_{\lambda}(s)$ is non-increasing in $\lambda$ for each $s \in \mathcal{S}.$
	

	\begin{remark}
		\begin{enumerate} 
			\item  It implies that for each $s \in \mathcal{S}$ there is $\lambda^*(s)$ such that  
			\begin{eqnarray*}
				\pi_{\lambda}(s) = 
				\begin{cases}
					1 & \mbox{$\lambda < \lambda^*(s), $} \\
					0 & \mbox{$\lambda \geq \lambda^*(s).$}
				\end{cases}
			\end{eqnarray*}
			Further, this suggests that as $\lambda$ increases, $B(\lambda)$ increases. Thus, a restless bandit is indexable.
			\item It is possible that a restless bandit can be indexable with the following structure on reward matrix.  $r(s,1)$ and $r(s,0)$ are non-decreasing in $s$ and there is no structure on transition probability matrices. Such a model is not considered in previous discussion.
			\item 
			It is further possible that a bandit is indexable even though there is no structure on transition probability and reward matrices. This is also not analyzed using previous Theorem.  
			This is due to limitation of  analysis. With this motivation, we study an alternative approach to indexability using value-iteration based methods in the next section.
			
		\end{enumerate}
	\end{remark}


\subsection{Alternative approach for indexability}

In this section, we study the value iteration algorithm and compute the optimal policy $\pi_{\lambda}(s)$ for each state $s \in \mathcal{S}$ for fixed value of subsidy $\lambda.$
We define the  grid of subsidy $\Lambda$ which is of finite size, say $J.$ We perform the value iteration for each $\lambda \in \Lambda$ and  compute the policy $\pi_{\lambda}(s)$ for $s \in \mathcal{S}.$ Size of policy matrix $\Phi = [[\pi_{\lambda}(s)]]$ is $K \times J.$  We next verify the indexability criteria in Remark~\ref{remark-1} using policy matrix $\Phi = [[\pi_{\lambda}(s)]],$  that is, there exist $\lambda^*$ such that 
\begin{eqnarray*}
	\pi_{\lambda}(s) = 
	\begin{cases} 
		1 & \mbox{ $\lambda < \lambda^*,$ } \\
		0 & \mbox{ $\lambda \geq \lambda^*,$}
	\end{cases}
\end{eqnarray*}
for each $ s \in \mathcal{S}.$  If this condition does not meet, we verify the non-indexable bandit condition in Remark~\ref{remark-1}. 
The value iteration algorithm for computation of policy matrix $\Phi$ is described in Algorithm~\ref{algo:VI}.

\begin{algorithm}[h]
	\KwIn{Reward matrix $R$, TP matrix $P,$ $\beta,$ tolerance $\Delta,$ Grid over subsidy $ \Lambda([-1,1]),$ $T_{\max}.$}
	\KwOut{Policy matrix $\Phi =[[\pi_{\lambda}(s)]]_{K \times J}$}
	\For{ $\lambda \in \Lambda([-1,1]) $} 
	{    
		Initialization: 
		$Q_0(s,a, \lambda) = 0$ for all $s \in \mathcal{S},$
		$a \in \{0,1\}.$ \\
		$V_0(s, \lambda) = 0$ for all $s \in \mathcal{S}.$\\
		
		\For {$s \in \mathcal{S}$}
		{
			\For {$t=0, 1, 2, \cdots, T_{\max}-1 $ }
			{
				\begin{eqnarray*}
					Q_{t+1}(s,a=0, \lambda) = r(s,a =0) + \lambda + \\ \beta \sum_{s^{\prime} \in S} p_{s,s^{\prime}}^0 V_t(s^{\prime},\lambda)
				\end{eqnarray*}
				\begin{eqnarray*}
					Q_{t+1}(s,a=1, \lambda) = r(s,a =1)  + \\ \beta \sum_{s^{\prime} \in S} p_{s,s^{\prime}}^1 V_t(s^{\prime},\lambda) 
				\end{eqnarray*}
				\begin{eqnarray*}
					V_{t+1}(s, \lambda) = \max \left\{ Q_{t+1}(s,a=0, \lambda) ,
					\right. \\ \left. 
					Q_{t+1}(s,a=1, \lambda)\right\} 
					\label{eqn:dp}
				\end{eqnarray*}	
			}
			
			\begin{eqnarray*}
				Q(s,a=0, \lambda) &=& Q_{T_{\max}}(s,a=0, \lambda) \\
				Q(s,a=1, \lambda) &=& Q_{T_{\max}}(s,a=1, \lambda)  
			\end{eqnarray*}
			\begin{eqnarray*} 
				V(s, \lambda) = \max\{  Q(s,a=0, \lambda),  Q(s,a=1, \lambda)\}  
			\end{eqnarray*} 
			
			\textbf{If} $\vert  Q(s,a=0, \lambda)-  Q(s,a=1, \lambda) \vert \geq \Delta$ \textbf{then} \\
			\begin{eqnarray*} 
				\Phi(s, \lambda) =\arg \max_{a \in \{0,1\}}  Q(s,a, \lambda)
			\end{eqnarray*} 
			
			\textbf{If} $\vert  Q(s,a=0, \lambda)-  Q(s,a=1, \lambda) \vert < \Delta$ \textbf{then} \\
			\begin{eqnarray*} 
				\Phi(s, \lambda) = 0
			\end{eqnarray*} 
		}
	}
	\Return{ Matrix $\Phi$} \;
	\caption{Value iteration algorithm for policy matrix $\Phi$ }
	\label{algo:VI}
\end{algorithm}

We now discuss the structure of matrix $\Phi$ with examples and this can depend on  transition probability and reward matrices.   

In the following examples we consider state space  $\mathcal{S} = \{1,2,3,4\}$ and $8$ values of   $\lambda \in \Lambda.$ Here $\Lambda = \{ \lambda_1, \lambda_2, \cdots, \lambda_8\}$ and $\lambda_i< \lambda_j$ for $1 \leq i<j \leq 8.$

\subsubsection{Example 1}  Suppose the policy matrix is as given in Eqn.~\eqref{example1}.

\begin{equation}
\Phi = 
\begin{blockarray}{ccccccccc}
\lambda_1 & \lambda_2 & \lambda_3 & \lambda_4 & \lambda_5 & \lambda_6 & \lambda_7 & \lambda_8 \\
\begin{block}{(cccccccc)c}
1 & 1 & 0 & 0 & 0 & 0 & 0 & 0 & s =1\\
1 & 1 & 1 & 0 & 0 & 0 & 0 & 0 & s =2\\
1 & 1 & 1 & 1 & 0 & 0 & 0 & 0 & s =3\\
1 & 1 & 1 & 1 & 1 & 1 & 0 & 0 & s =4\\
\end{block}
\end{blockarray}
\label{example1}
\end{equation}

Observe that for fixed $\lambda,$ $\pi_{\lambda}= \Phi(:,\lambda)$ is the policy for state vector. The policy $\pi_{\lambda}(s)$ is non-decreasing in $s.$ This structure implies a single threshold policy in $s$ for fixed subsidy $\lambda.$ Moreover for fixed $s,$ as $\lambda$ increases then $\pi_{\lambda}(s)$ is non-increasing in $\lambda.$ This structure implies that there is a single threshold policy in $\lambda$ for fixed values of state $s.$
If a single threshold policy in $\lambda$ exists for each $s \in \mathcal{S},$ then we can say that bandit is indexable, and  this follows from the definition of indexability because $B(\lambda)$ is a  non-decreasing set  as we increase $\lambda.$ 
In this example, $B(\lambda_1) = \emptyset$, $B(\lambda_2)= \emptyset$, $B(\lambda_3) = \{1\}$, $B(\lambda_4) = \{1,2\}$, $B(\lambda_5) = \{1,2,3\}$, $B(\lambda_6) = \{1,2,3\}$, $B(\lambda_7) = \{1,2,3,4\}$, and $B(\lambda_8) = \{1,2,3,4\}$.

\subsubsection{Example 2} Suppose the policy matrix is as given in Eqn.~\eqref{example2}.

\begin{equation}
\Phi = 
\begin{blockarray}{ccccccccc}
\lambda_1 & \lambda_2 & \lambda_3 & \lambda_4 & \lambda_5 & \lambda_6 & \lambda_7 & \lambda_8 \\
\begin{block}{(cccccccc)c}
1 & 1 & 0 & 0 & 0 & 0 & 0 & 0 & s =1\\
1 & 1 & 1 & 1 & 1 & 0 & 0 & 0 & s =2\\
1 & 1 & 1 & 0 & 0 & 0 & 0 & 0 & s =3\\
1 & 1 & 1 & 1 & 1 & 1 & 0 & 0 & s =4\\
\end{block}
\end{blockarray}
\label{example2}
\end{equation}

Notice that for fixed $\lambda,$ $\pi_{\lambda}= \Phi(:,\lambda)$ and the policy $\pi_{\lambda}(s)$ is not monotone in $s$ for all $\lambda \in  \Lambda.$  We can observe from~\eqref{example2} that for $\lambda_4,$ and $\lambda_5$ the optimal policy $\pi_4 = \pi_5 = [0,1,0,1]^T$ and there is no single threshold policy in $s.$ 

But for fixed $s,$ as $\lambda$ increases then $\pi_{\lambda}(s)$ is non-increasing in $\lambda.$ This structure implies that there is a single threshold policy in $\lambda$ for fixed values of state $s \in \mathcal{S}.$ 
$B(\lambda)$ is a non-decreasing set as we increase $\lambda.$ 
In this example, $B(\lambda_1) = \emptyset$, $B(\lambda_2)= \emptyset$, $B(\lambda_3) = \{1\}$, $B(\lambda_4) = \{1,3\}$, $B(\lambda_5) = \{1,3\}$, $B(\lambda_6) = \{1,2,3\}$, $B(\lambda_7) = \{1,2,3,4\}$, and $B(\lambda_8) = \{1,2,3,4\}$.
From the definition of indexability, we can say that the bandit is indexable.

\subsubsection{Example 3} Suppose the policy matrix is as given in Eqn.~\eqref{example3}.

\begin{equation}
\Phi = 
\begin{blockarray}{ccccccccc}
\lambda_1 & \lambda_2 & \lambda_3 & \lambda_4 & \lambda_5 & \lambda_6 & \lambda_7 & \lambda_8 \\
\begin{block}{(cccccccc)c}
1 & 1 & 0 & 0 & 0 & 0 & 0 & 0 & s =1\\
1 & 1 & 1 & 1 & 1 & 0 & 0 & 0 & s =2\\
1 & 1 & 1 & 0 & 0 & 1 & 1 & 0 & s =3\\
1 & 1 & 1 & 1 & 1 & 1 & 0 & 0 & s =4\\
\end{block}
\end{blockarray}
\label{example3}
\end{equation}

In this example, $B(\lambda_1) = \emptyset$, $B(\lambda_2)= \emptyset$, $B(\lambda_3) = \{1\}$, $B(\lambda_4) = \{1,3\}$, $B(\lambda_5) = \{1,3\}$, $B(\lambda_6) = \{1,2\}$, $B(\lambda_7) = \{1,2,4\}$, and $B(\lambda_8) = \{1,2,3,4\}$.  
Observe that $B(\lambda_{5}) \not\subset B(\lambda_{6})$ for $\lambda_5 < \lambda_6.$ Hence, the bandit is not indexable. 

Notice that for fixed $\lambda,$ $\pi_{\lambda}= \Phi(:,\lambda)$ and the policy $\pi_{\lambda}(s)$ is not monotone in $s$ for all $\lambda \in  \Lambda.$  There is no single threshold policy in $s$ for $\lambda_4, \lambda_5,$ and $ \lambda_7.$

For fixed $s =3,$ as $\lambda$ increases, then $\pi_{\lambda}(s)$ is not monotone in $\lambda,$ because policy is $\{1,1,1,0,0,1,1,0\}.$ This structure implies that there is no single threshold policy in $\lambda$  for $s=3.$ Because of this, it does not satisfy condition of indexability. 


\subsection{Computational complexity of Verification of Indexability and Index Computation} 

RMABs are known to be a PSPACE hard problem,\cite[Theorem $4$]{Papadimitriou1999}. However, the Lagrangian relaxed  RMAB problem is solved by the index policy, where we analyze a single-armed restless bandit. Each bandit is a MDP with finite number of states and two actions. From \cite{Papadimitriou1987}, MDP with finite state-space and finite action space can be solved in polynomial in time (P-complete). Thus, the computational complexity of indexability and index computation is polynomial in time. Note that we have used the value iteration algorithm  for computation of policy matrix with fixed grid size of subsidy, $|\Lambda| = J.$  From \cite{Littman1995},  the sample complexity of value iteration algorithm is  $O\left( \frac{K^2\log(1/(1-\beta)\epsilon)}{1-\beta}\right).$ 
The value iteration algorithm is performed for each $\lambda \in \Lambda.$ 
Thus, the sample complexity of computing policy matrix $\Phi$ is $O\left( \frac{K^2 J \log(1/(1-\beta)\epsilon)}{1-\beta}\right).$ The verification of indexability from matrix $\Phi$ requires additional time $KJ.$ This  is due to the fact that one has to verify for each $s,$ $\Phi(s, \lambda)$ is non-increasing in $\lambda$ for all $\lambda \in \Lambda.$ Hence, indexability verification criteria and the index computation require sample complexity of $O\left( KJ+ \frac{K^2 J \log(1/(1-\beta)\epsilon)}{1-\beta}\right).$ 
There are many variants of  value iteration algorithm which have improved the sample complexity in terms of its dependence on number of states $K,$ see \cite{Wang2019}. Therefore, one can improve the complexities of indexability verification and index computation.   

\section{Algorithms for restless multi-armed bandits}
\label{sec:algo-rmab}
We now discuss algorithms for restless multi-armed bandits---myopic policy, Whittle index policy and online rollout policy.  In the myopic policy, the arms with the highest immediate rewards are selected in each time-step. That is,  $(X_t^{1}, X_t^{2}, \cdots, X_t^{N}) \in \mathcal{S}^{N},$ then the immediate reward for arm $n$ is $r^{n}(X_t^n, A_t^n).$ Thus, $M$ arms are selected using immediate reward at each time-step $t.$ In the Whittle index policy, for given time-step $t$ with state $(X_t^{1}, X_t^{2}, \cdots, X_t^{N}) \in \mathcal{S}^{N},$ one computes the index for each bandit, i.e.,  the index for $n$th bandit is $W(X_t^n).$ Then, we select the $M$ arms with the highest Whittle indices. We compute the Whittle indices for all states in the offline mode. In the following we discuss online rollout policy.

\subsection{Online Rollout Policy}
\label{sec:online-rollout-policy}
We now discuss a simulation-based approach referred to as online rollout policy. Here, many trajectories are `rolled out' using a simulator and the value of each action is estimated based on the cumulative reward along these trajectories. 
Trajectories of length $H$ are generated using a fixed base policy, say, $\phi,$ which might choose arms according to a deterministic rule (for instance, myopic decision) at each time-step. The information obtained from a trajectory is 
\begin{eqnarray} 
\{ X^{n}_{h,l},  b^{n,\phi}_{h,l}, r^{n,\phi}_{h,l} \}_{n=1,h=1}^{ N, H}
\end{eqnarray} 
under policy $\phi.$ Here, $l$ denotes a trajectory, $h$ denotes time-step and $X^{n}_{h,l}$ denotes the state about arm $n.$ The action of playing or not playing arm $n$ at time-step $h$ in trajectory $l$ is denoted by $b^{n,\phi}_{h,l} \in \{0,1\}.$ Reward obtained from arm $n$ is $ r^{n,\phi}_{h,l}.$ 

We now describe the rollout policy for $M=1$. 	
We compute the value estimate for trajectory $l$ with starting belief $X = (X^{1}, \cdots, X^N),$  and initial action $\xi\in \{1,2, \cdots, N\}$.  Here, $\xi_{h,l} = n$ means arm $n$ is played at time-step $h$ in trajectory $l$, so $b^{n,\phi}_{h,l} = 1$, and  $b^{i,\phi}_{h,l} = 0$ for $\forall i \neq n$.
The value estimate for initial action $\xi$ along trajectory $l$ is given by 
\begin{eqnarray*}
	Q_{l}^{\phi}( X, \xi ) &=& \sum_{h=1}^{H} \beta^{h-1} \sum_{n=1}^{N} r^{n,\phi}_{h,l}.
\end{eqnarray*} 
Then, averaging over $L$ trajectories, the value estimate for action $\xi$ in state $X$    under policy $\phi$ is 
\begin{eqnarray*}
	\widetilde{Q}_{H,L}^{\phi}({X,  \xi}) = \frac{1}{L}\sum_{l=1}^{L}  Q_{l}^{\phi}( X, \xi ) ,
\end{eqnarray*} 
where, the base policy $\phi$ is myopic (greedy), i.e., it chooses the arm with the highest immediate reward, along each trajectory. Now, we perform one-step policy improvement, and the optimal action is selected as, 
\begin{align}
j^*(X) = \arg \max_{1 \leq j \leq N} \left[ \widetilde{r}(X, \xi = j) + \beta \widetilde{Q}_{H,L}^{\phi}(X,\xi=j) \right], 
\label{eqn:policy-improv}
\end{align}
and $\widetilde{r}(X, \xi ) = \sum_{n=1}^{N} r^n(X^n,  b^n ).$ 
In each time-step $t$  with $X_t = \{X^1_t, \cdots, X^n_t \},$ online rollout policy plays the arm $j^*(X_t)$ obtained according to Eqn.~\eqref{eqn:policy-improv}.  This can be extended to multiple play of arms.


We now discuss the computational complexity of online rollout policy. As rollout policy is a heuristic (look-ahead)  policy, it does not require convergence analysis. In the Whittle index policy, index computation is done offline, where we compute and store the index values for each element on the grid $G,$ for all arms ($N$). 
During online implementation, when state $X = \{ X^1,...,X^N \}$ is observed, the corresponding index values are drawn from the stored data. On the other hand,  rollout policy is implemented online and its  computational complexity is stated in the following Theorem. 
\begin{theorem}
	The online rollout policy has a worst case complexity of $O(N(HL+2)T)$ for number of iterations $T,$ when the base policy is myopic. 
	\label{thm:complexity-rollout}
\end{theorem}
The proof is given in Appendix  \ref{app:sec-thm-complexity-rollout}. Here $T$ is dependent on $\beta$ and $T = 1/(1-\beta).$

\subsection{Comparison of complexity results}
For large state space, $K = 1000,$ one requires grid  of subsidy $\Lambda$ to be large, that is, $J = O(K).$ Thus the sample complexity of index computation is $O\left( K^2 + \frac{K^3 \log(1/(1-\beta)\epsilon)}{1-\beta}\right).$ The computational complexity of online rollout policy is dependent on horizon length $H$ and number of trajectories $L$ and it is $O(N(HL+2)/(1-
\beta)).$ For large state space model, rollout policy can have lower computational complexity compared to index policy.

\section{Numerical Examples of single-armed restless bandit}
\label{sec:numerical-indexability-sab}
In this section, we provide numerical examples to illustrate the indexability of the bandit using our approach. In all following examples, we use discount parameter $\beta = 0.9.$ Similar results are also true for $\beta = 0.99.$
\subsection{Example with circular dynamics} 

In the following example, we illustrate  that the optimal policy is not a single threshold type  but  it is indexable. This example is taken from \cite{Avrachenkov2022}, \cite{Fu2019}, where  indexability of the bandit is assumed, and author  studied learning algorithm. In this example, we claim indexability using our approach.  

The example has four states and the dynamics are circulant: when an arm is passive ($a = 0$), resp. active ($a = 1$), the state evolves according to the transition probability matrices. 
\begin{eqnarray*} 
	P^0 = 
	\begin{bmatrix}
		1/2 & 0 & 0 & 1/2 \\
		1/2 & 1/2 & 0 & 0 \\
		0 & 1/2 & 1/2 & 0 \\
		0 & 0 & 1/2 & 1/2
	\end{bmatrix},
\end{eqnarray*}
\begin{eqnarray*} 
	P^1 = {P^0}^T.
\end{eqnarray*}
The reward matrix is given as follows.
\begin{eqnarray*} 
	R = 
	\begin{bmatrix}
		-1 & -1 \\
		0 & 0 \\
		0 & 0  \\
		1 & 1
	\end{bmatrix}.
\end{eqnarray*}

\begin{figure}[h] 
	\begin{center} 
		\includegraphics[scale=0.2]{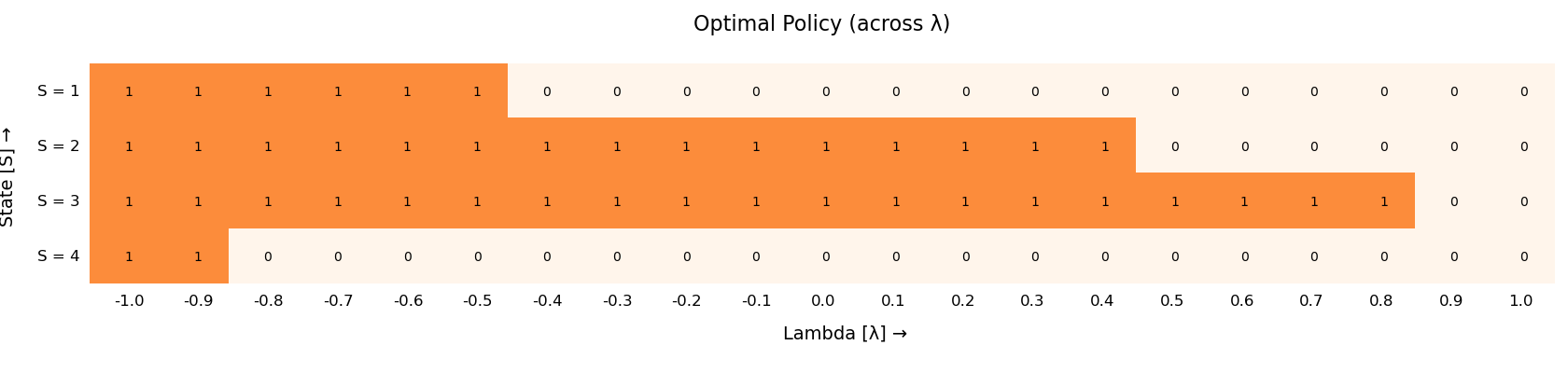}
	\end{center}
	\caption{Example of circular dynamics: policy matrix $\Phi$ for different states as function of subsidy $\lambda.$}
	\label{fig:circulant-model}
\end{figure}

We consider different values of subsidy $\lambda \in [-1,1]$  and compute the  optimal policy for $s \in \mathcal{S}$ using dynamic programming algorithm and obtain the policy matrix $\Phi,$ which is given in Fig.~\ref{fig:circulant-model}. We observe  that the optimal policy has more than two threshold for some values of subsidy $\lambda.$  This is clear for  subsidy $\lambda = -0.4,$ the optimal policy $\pi_{\lambda = -0.4} = \{0,1,1,0\}$ and there are two  switches in decision rule. But it is indexable because as $\lambda$ increases from $-1$  to $+1,$ then the set $B(\lambda)$ is non-decreasing, where $B(\lambda) = \{ s \in \mathcal{S}: \pi_{\lambda}(s) = 0 \}. $ Note  that $B(-0.9) = \emptyset,$ $B(-0.8) = \{4\},$ $B(-0.4) = \{ 1,4\},$ $B(0.5) = \{1,2,4\},$  $B(0.9) = \{1,2,3,4\}.$ 


\subsection{Examples with restart}
We illustrate the structure of policy matrix $\Phi$ for restart model, where action $1$ is taken, then $p^{1}_{s,1} = 1$ for all $s \in \mathcal{S}.$  
We consider  $K=5$ states.
\begin{figure}[h] 
	\begin{center} 
		\includegraphics[scale=0.3]{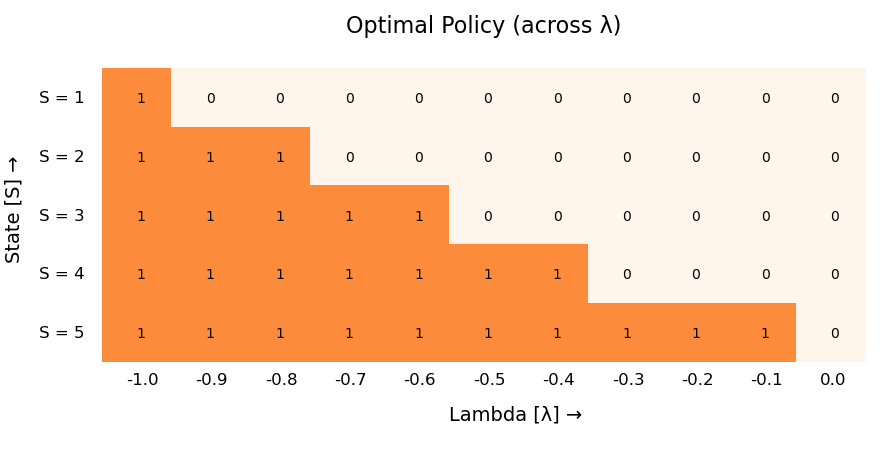}
	\end{center} 
	\caption{Example of restart model: policy matrix $\Phi$ for different states as function of subsidy $\lambda.$ States $K=5.$}
	\label{fig:restart-model-5states}
\end{figure} 
This example is taken from \cite{Avrachenkov2022}, where authors assumed indexability. The transition probability matrices are as follows. 

\begin{eqnarray*}
	P_0 &=&  \begin{bmatrix}
		1/10 & 9/10 & 0 & 0 & 0 \\
		1/10 & 0 & 9/10 & 0 & 0 \\
		1/10 & 0 & 0 & 9/10 & 0 \\
		1/10 & 0 & 0 & 0 & 9/10 \\
		1/10 & 0 & 0 & 0 & 9/10
	\end{bmatrix}, \\
	P_1 &=& 
	\begin{bmatrix}
		1 & 0 & 0 & 0 & 0 \\
		1 & 0 & 0 & 0 & 0 \\
		1 & 0 & 0 & 0 & 0 \\
		1 & 0 & 0 & 0 & 0 \\
		1 & 0 & 0 & 0 & 0
	\end{bmatrix}.
\end{eqnarray*}
The rewards for  passive action $(a=0)$ with state $k,$ $r(k,0) = 0.9^k$ and the rewards for active action  $(a=1)$, $r(k,1) =0$ for all $k \in \mathcal{S}.$ 
We observe from Fig.~\ref{fig:restart-model-5states} that the optimal policy is of a single threshold type for fixed subsidy $\lambda.$ Moreover, as subsidy increases from $-1$ to $+1,$ an optimal policy  changes monotonically  from active action to passive action for all states. Thus, the bandit is indexable. 

\subsection{Non-Indexable model with $5$ states} 
\begin{figure}[h] 
	\begin{center} 
		\includegraphics[scale=0.2]{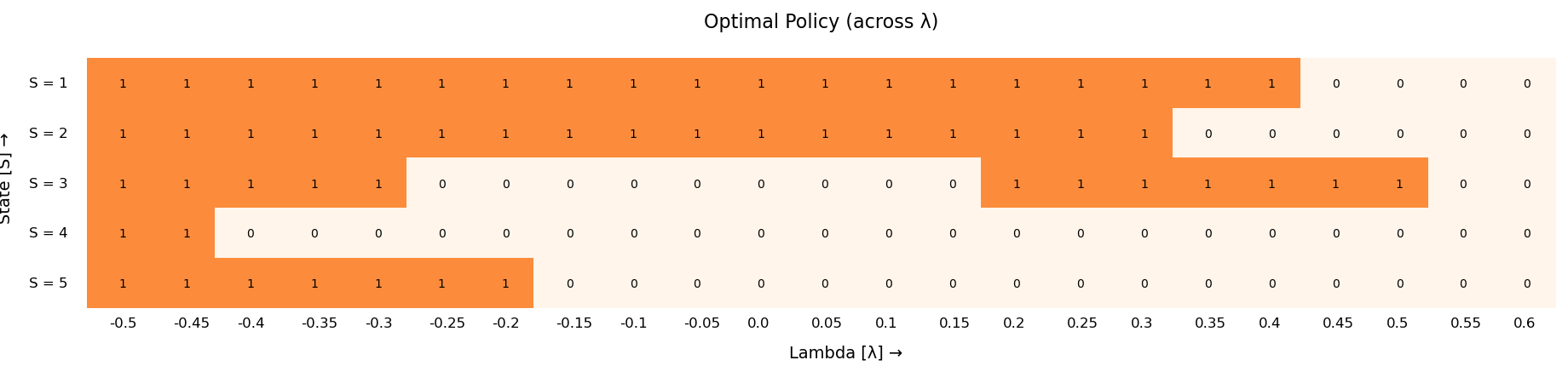}
	\end{center}
	\caption{Example of a model with no structure on  prob. matrices : policy matrix $\Phi$ for different states as function of subsidy $\lambda.$ States $K=5.$ Non-indexable example.}
	\label{fig:Our-model-5state-2action-Non-indexable}
\end{figure}

We consider following parameters.
\begin{eqnarray*}
	P_0 &=&  \begin{bmatrix}
		0.1502 & 0.0400 & 0.4156 & 0.0300 & 0.3642 \\
		0.4000 & 0.3500& 0.0800& 0.1200& 0.0500 \\
		0.5276 & 0.0400 &0.3991 & 0.0200 & 0.0133  \\
		0.0500 & 0.1000& 0.1500& 0.2000& 0.5000 \\
		0.0191 & 0.0100 & 0.0897 & 0.0300 & 0.8512 
	\end{bmatrix}, \\
	P_1 &=& 
	\begin{bmatrix}
		0.7196 & 0.0500 & 0.0903 & 0.0100 & 0.1301 \\
		0.5500 & 0.2000& 0.0500& 0.0800& 0.1200 \\ 
		0.1903 & 0.0100 & 0.1663 & 0.0100 & 0.6234 \\
		0.2000 & 0.0500 &0.1500 & 0.1000 & 0.5000 \\
		0.2501 & 0.0100& 0.3901 & 0.0300 & 0.3198 
	\end{bmatrix}, \\
	R &= &
	\begin{bmatrix}
		0.4580 & 0.9631  \\
		0.5100 & 0.8100 \\
		0.5308 & 0.7963  \\
		0.6710 & 0.1061 \\
		0.6873 & 0.1057  
	\end{bmatrix}.
\end{eqnarray*}
We make no structural assumption on transition probability matrices. But the reward is decreasing in $s$ for  $a=1$ and increasing in $s$ for $a=0.$ From Fig.~\ref{fig:Our-model-5state-2action-Non-indexable}, we notice that $\pi_{\lambda}(s)$ is not monotone in $\lambda$ for $s=3.$  Thus,  the bandit is non-indexable.

More numerical examples on indexability  are provided in 
Appendix.

\section{Numerical results on RMAB}
\label{sec;numerical-example-rmab}
\begin{figure}[h] 
	\begin{center} 
		\includegraphics[scale=0.37]{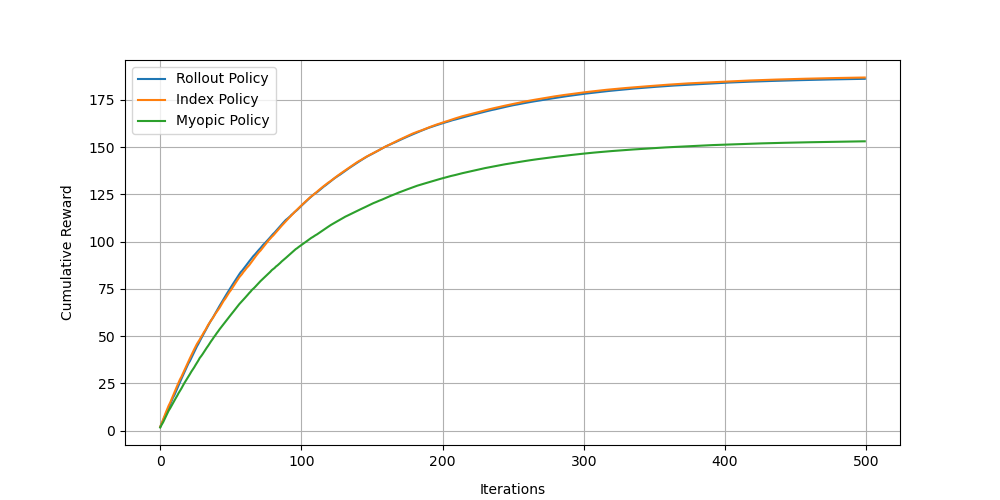}
	\end{center} 
	\caption{$3$ armed restless bandits: Non-identical, monotone and indexable}
	\label{fig:RMAB-1}
\end{figure}

\begin{figure}[h] 
	\begin{center} 
		\includegraphics[scale=0.37]{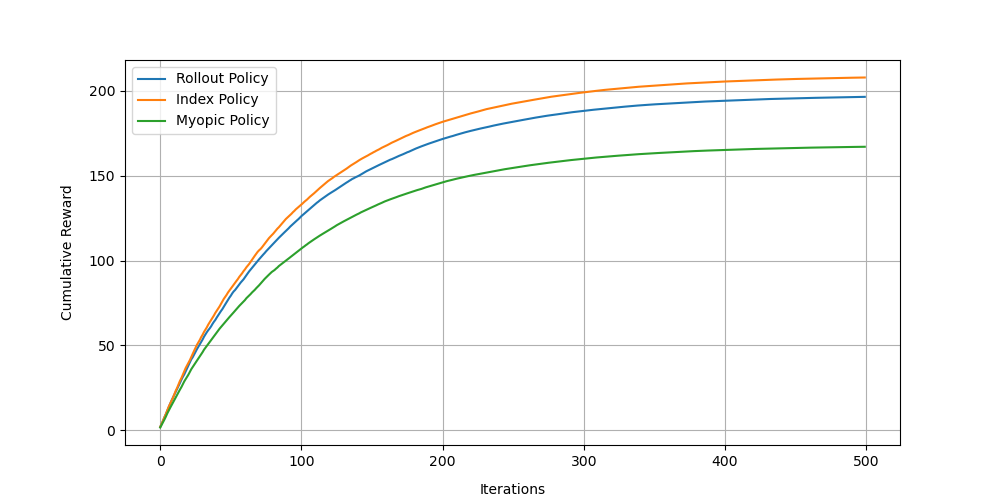}
	\end{center} 
	\caption{$5$ armed restless bandits: Non-identical, monotone and indexable}
	\label{fig:RMAB-2}
\end{figure}

\begin{figure}[h] 
	\begin{center} 
		\includegraphics[scale=0.37]{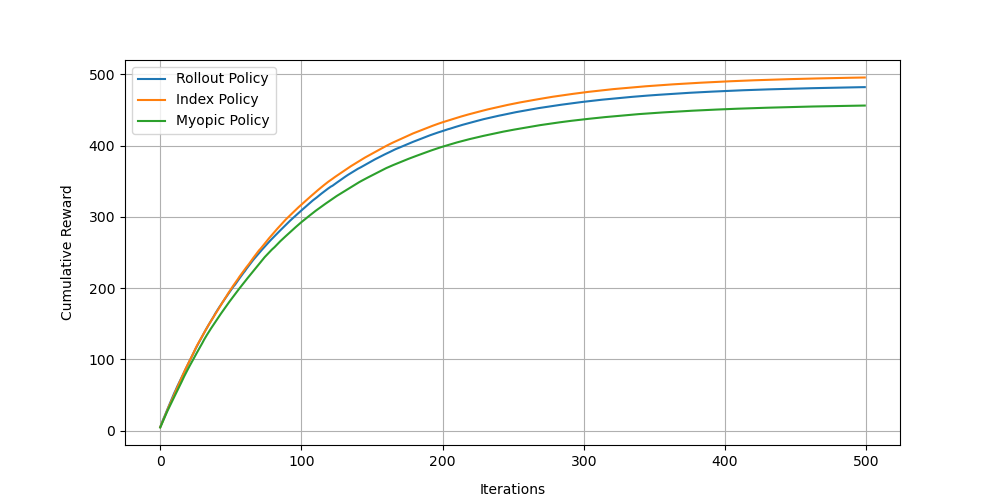}
	\end{center} 
	\caption{$10$ armed restless bandits: Non-identical, monotone and indexable}
	\label{fig:RMAB-3}
\end{figure}

We compare the performance of rollout policy, index policy and myopic policy. 
We consider  number of arms $N=3,5,10,$ and discount parameter $\beta =0.99.$   In the rollout policy, we used number of trajectories $L=30,$ and horizon length $H=4.$
In the following, we consider non-identical restless bandits. All bandits are  indexable, and index is monotone in state. In Fig.~\ref{fig:RMAB-1}, ~\ref{fig:RMAB-2} and ~\ref{fig:RMAB-3}, we compare performance of myopic, rollout and index policy for $N=3,5,10,$ respectively. We observe that the rollout policy and the index policy performs better than myopic policy. 

Additional numerical examples of RMAB are given in  Appendix. 

\section{Discussion and concluding remarks} 
\label{sec:concluding-remark}
In this paper, we studied finite state restless bandits problem. We analyzed the Whittle index policy and proposed a value-iteration based algorithm, which computes the policy matrix, verifies indexability condition and obtains the indices. Further, the sample complexity analysis is discussed for index computation. We also study online rollout policy and its computational complexity. Numerical examples are provided to illustrate the indexability condition and performance of algorithms---myopic policy, index policy and rollout policy. 

From numerical examples, we note that the indexability condition does not require any structural assumption on transition probabilities. 
It is even  possible that the arms having rewards with no structural assumption with respect to actions can be indexable. Hence, our approach can be widely applicable without any structural model assumptions to verify the indexability condition and index computation. In future work, we plan to extend our approach to countable state-space model.

\section*{Acknowledgment} 
R. Meshram and S.~Prakash are with  ECE Dept. IIIT Allahabad and Center for Intelligent Robotics (CIR) Lab, IIIT Allahabad, India. 
V.~Mittal and D.~Dev are with ECE Dept. IIIT Allahabad, India.
The work of Rahul Meshram and Deepak Dev is supported by grants from SERB.

\bibliographystyle{IEEEbib}

\bibliography{its-references}

\newpage

\appendix

\subsection{Proof of Lemma~\ref{lemma:convex}}
\label{app:lemma:convex}

Proof is by induction method. We provide the proof for piecewise convex. 

Let 
\begin{eqnarray*}
	Q_0(s,a=1,\lambda) &=& r(s,1) \\ 
	Q_0(s,a=0,\lambda) &=& r(s,0) + \lambda \\
	V_1(s, \lambda)  &=& \max \{ r(s,1), r(s,0)+\lambda \}    
\end{eqnarray*} 
Note that rewards $r(s,1)$ and $r(s,0)$ are linear functions. The max of linear functions is piecewise convex function. 

We next assume that $V_{n}(s,\lambda)$ is piecewise convex in $s$ for fixed $\lambda, \beta.$ 

From dynamic programming equations, we have 
\begin{eqnarray*}
	Q_{n}(s,a,\lambda) = r(s,a)+\lambda(1-a) + \beta \sum_{s^{\prime} \in \mathcal{S} } p_{s,s^{\prime}}^a V_n(s^{\prime},\lambda).
\end{eqnarray*}
We want to show that $Q_{n}(s,a,\lambda)$ is piecewise convex in $s.$ 
Now observe that 
$\sum_{s^{\prime} \in \mathcal{S} } p_{s,s^{\prime}}^0 V_n(s^{\prime},\lambda)$ is 
piecewise convex in $s$ because it is a convex combination of piecewise convex functions. 
Hence $Q_{n}(s,a,\lambda)$ is piecewise convex function in $s.$ Then 
\begin{eqnarray*}
	V_{n+1}(s, \lambda)  &=& \max \{Q_{n}(s,0,\lambda), Q_{n}(s,1,\lambda) \} 
\end{eqnarray*}
$V_{n+1}(s)$ is piecewise convex in $s.$ Hence $V_n(s)$ is piecewise convex in $s$ for all $n.$

From Banach fixed point theorem, \cite[Theorem $6.2.3$]{Puterman2005}, the dynamic program equation converges and there exists unique $V.$ Thus 
as $n \rightarrow \infty, $ $V_{n}(s,\lambda) \rightarrow V(s,\lambda)$  and hence $V(s,\lambda)$ is piecewise convex in $s.$ The result follows.

Analogously, we can prove that $V(s,\lambda)$ piecewise convex in $\lambda$ for fixed $s, \beta.$

\subsection{Proof of Theorem~\ref{thm:monotone-s}}

We make use of following Lemma from \cite[Lemma $4.7.2$]{Puterman2005}.

\begin{lemma}
	Let $\{y_j\}$ and $\{y_j^{\prime}\}$ be real-valued non-negative sequences satisfying 
	\begin{eqnarray*}
		\sum_{j=k} y_j \geq \sum_{j=k} y_j^{\prime}
	\end{eqnarray*}
	for all values of $k.$ Equality holds for $k=0.$
	Suppose that $v_{j+1} \geq  v_j$ for $j =0,1,2, \cdots,$ then 
	\begin{eqnarray}
	\sum_{j=k} v_j y_j \geq  \sum_{j=k} v_j y_j^{\prime}.
	\end{eqnarray}
\end{lemma}
Using lemma we have following results. 
Let $Z$ and $Z^{\prime}$  be two random variables with probabilities $\prob{Z = j} = y_j$ and $\prob{Z^{\prime} = j} = y_j^{\prime}.$ If $\sum_{j=k} y_j \geq \sum_{j=k} y_j^{\prime},$ then $Z$ is  stochastically greater than  $Z^{\prime}.$ From preceding lemma for any non-decreasing function $f(j)$ we have $\expect{f(Z)} \geq \expect{f(Z^{\prime})}.$
We first prove the following lemma.

\begin{lemma}
	Suppose 
	\begin{enumerate}
		\item $r(s,a)$ is non-decreasing in $s$ for all $a \in \{0,1\},$
		\item $q(k|s,a) = \sum_{j=k}^{K}p_{s,j}^a$ is non-decreasing in $s.$
	\end{enumerate}
	Then $V(s, \lambda)$ is non-decreasing in $s$ for fixed $\lambda.$
\end{lemma}

\begin{IEEEproof}
	Proof of this lemma is by induction method. 
	Let 
	\begin{eqnarray*}
		Q_0(s,a=1,\lambda) &=& r(s,1) \\ 
		Q_0(s,a=0,\lambda) &=& r(s,0) + \lambda \\
		V_1(s, \lambda)  &=& \max \{ r(s,1), r(s,0)+\lambda \}    
	\end{eqnarray*} 
	Notice  that $Q_0(s,a=1,\lambda),$  $Q_0(s,a=0,\lambda)$ are non-decreasing in $s,$ and hence $V_1(s, \lambda) $ is non-decreasing in $s.$ 
	
	Assume that $V_{n}(s,\lambda)$ is non-decreasing in $s$ for fixed $\lambda.$ 
	Then,
	\begin{eqnarray*}
		Q_{n}(s,a,\lambda) = r(s,a)+\lambda (1-a)+ \beta \sum_{s^{\prime} \in \mathcal{S} } p_{s,s^{\prime}}^a V_n(s^{\prime},\lambda).
	\end{eqnarray*}
	
	From earlier lemma and discussion,  $\sum_{s^{\prime} \in \mathcal{S} } p_{s,s^{\prime}}^0 V_n(s^{\prime},\lambda)$ is non-decreasing in $s.$ whenever $V_n(s,\lambda)$ is non-decreasing in $s.$
	Hence $Q_{n}(s,0,\lambda)$ is non-decreasing in $s.$
	Similarly, $Q_{n}(s,1,\lambda)$ is non-decreasing in $s.$ Then 
	\begin{eqnarray*}
		V_{n+1}(s, \lambda)  &=& \max \{Q_{n}(s,0,\lambda), Q_{n}(s,1,\lambda) \} 
	\end{eqnarray*}
	and  $V_{n+1}(s)$ is non-decreasing in $s.$ Hence $V_n(s)$ is non-decreasing in $s$ for all $n.$
	
	From Banach fixed point theorem, \cite[Theorem $6.2.3$]{Puterman2005}, the dynamic program equation converges and there exists unique $V.$ Thus 
	as $n \rightarrow \infty, $ $V_{n}(s,\lambda) \rightarrow V(s,\lambda)$  and hence $V(s,\lambda)$ is non-decreasing in $s.$ The result follows. 
\end{IEEEproof} 

\begin{lemma}
	$V(s, \lambda)$ is non-decreasing in $\lambda$ for fixed $s.$
\end{lemma}
The proof of this is again by induction method, proof steps are analogous to previous lemma and hence we omit the details of the proof. 

We are now ready to prove the main  theorem. 

\begin{enumerate}
	\item If $r(s,a)$ is a superadditive function on $\mathcal{S} \times \{0,1\},$ then  
	\begin{eqnarray*}
		r(s^{\prime},1) -r(s^{\prime},0)  \geq r(s,1) -r(s,0)
	\end{eqnarray*}
	for $s^{\prime} > s$ and $s^{\prime},s \in \mathcal{S}.$
	\item  If $q(k|s,a)$ is a superadditive function on $\mathcal{S} \times \{0,1\},$ then 
	\begin{eqnarray*}
		q(k|s^{\prime},1) - q(k|s^{\prime},0)  \geq  q(k|s,1)  -  q(k|s,0)
	\end{eqnarray*}
	for all $k\in \mathcal{S}.$
	This implies  
	\begin{eqnarray*}
		\sum_{j=k}^{K} \left[ p_{s^{\prime},j}^{1} - p_{s^{\prime},j}^{0 } \right] \geq 
		\sum_{j=k}^{K} \left[ p_{s,j}^{1} - p_{s,j}^{0 } \right].
	\end{eqnarray*}
\end{enumerate} 
We assumed that  $\lambda$ is fixed.
Proof is by induction method. 
From assumption on rewards, $Q_0(s,a,\lambda)$ is superadditive on $\mathcal{S}\times\{0,1\}.$ 
The optimal policy is 
\begin{eqnarray}
\pi_{\lambda,1}(s) = \arg\max_{a \{0,1\}} \{ Q_0(s,a,\lambda),\},
\end{eqnarray}
Further, note that 
\begin{eqnarray*}Q_0(s,1,\lambda) - Q_0(s,0,\lambda) = r(s,1) - r(s,0) -\lambda
\end{eqnarray*}
and $Q_0(s,1,\lambda) - Q_0(s,0,\lambda)$ is non-decreasing in $s$ for fixed $\lambda.$
Hence, the optimal policy $\pi_{\lambda,1}(s)$ is non-decreasing in $s.$

We next want to show that the optimal policy $\pi_{\lambda,n+1}$ at time step $n+1$ is non-decreasing in $s.$ 
We need to show that $Q_n(s,1,\lambda) - Q_n(s,0,\lambda)$
is non-decreasing in $s.$

Consider 
\begin{eqnarray*}
	Q_n(s,1,\lambda) - Q_n(s,0,\lambda) 
	= r(s,1) -r(s,0) - \lambda + \\ 
	\beta 
	\left[ \sum_{j=k}^{K} (p_{s,j}^1 - p_{s,j}^0) V_n(j,\lambda) \right].
\end{eqnarray*}

Next we require the following inequality for $s^{\prime} \geq s$ to be true.
\begin{eqnarray*}
	\sum_{j=k}^{K} \left[ p_{s^{\prime},j}^{1} - p_{s^{\prime},j}^{0 } \right] V_n(j,\lambda) \geq 
	\sum_{j=k}^{K} \left[ p_{s,j}^{1} - p_{s,j}^{0 } \right] V_n(j,\lambda).
\end{eqnarray*}
From earlier lemma, we know that $V_n(j)$ is non-decreasing in $s$ and $q(k|s,a)$ is a superadditive function on $\mathcal{S} \times \{0,1\}.$ 
Thus $\sum_{j=0}^{K} p_{s,j}^aV_n(j,\lambda)$  is superadditive.  
Reward $r(s,a)$ is a superadditive function on $\mathcal{S} \times \{0,1\},$
Hence $Q_{n}(s,a,\lambda) $ is a superadditive function on $\mathcal{S} \times \{0,1\}$ for all $n.$ 

This implies that  $Q_n(s,1,\lambda) - Q_n(s,0,\lambda)$
is non-decreasing in $s$ and there exists the  optimal policy $\pi_{\lambda,n+1}(s) = \arg\max_{a} Q_n(s,a,\lambda)$ which is non-decreasing in $s$ for fixed $\lambda.$ 
The policy $\pi_{\lambda,n}(s)$ is non-decreasing in $s$ for all values of $n.$

From Banach fixed point theorem, \cite[Theorem $6.2.3$]{Puterman2005}, the dynamic program equation converges and there exists unique $V.$ Thus 
as $n \rightarrow \infty, $ $V_{n}(s,\lambda) \rightarrow V(s,\lambda)$
and $V(s,\lambda)$ is non-decreasing in $s,$ 
$\pi_{\lambda, n}(s) \rightarrow \pi_{\lambda}(s).$
Moreover, 
\begin{eqnarray*}
	Q(s,a, \lambda) = 
	r(s,a) + \lambda (1-a) + \beta \sum_{j=1}^{K} p_{s,j}^a V(j, \lambda).
\end{eqnarray*}
Then $Q(s,1,\lambda) - Q(s,0,\lambda)$ is non-decreasing in $s.$ Hence $\pi_{\lambda}(s)$ is non-decreasing in $s$ for fixed $\lambda.$
This completes the proof. 

\subsection{Proof of Theorem~\ref{thm:complexity-rollout}}
\label{app:sec-thm-complexity-rollout}

In \cite{Meshram22},  complexity analysis is studied for online rollout policy with partially observable restless bandits. Here, we sketch proof detail with finite state observable restless bandit model.

We consider  case  $M=1,$ i.e., a single-arm is played.

Recall from Section~\ref{sec:online-rollout-policy} that we simulate the multiple trajectories for fixed horizon length. We compute the value estimate for each trajectory, say $l$ starting from  initial state $X,$ and initial action $\xi,$
i.e., $Q_{l}^{\phi}(X, \xi) .$ This require computation of $O(H),$ since each trajectory is run for $H$ horizon length. There are $L$ trajectories and it requires $O(HL)$ computation.

The empirical value estimates $\widetilde{Q}_{H,L}^{\phi}(X, \xi),$ $\xi \in \{1,2,\cdots, N\}$ for $N$ possible initial actions (arms). This takes $O(NHL)$ computations as there are $L$ trajectories of horizon (look-ahead) length $H$ for each of the $N$ initial actions. 

Rollout policy also involves the policy improvement step and it takes another $O(2N)$ computations. Thus total computation complexity in each iteration is $O(2N + NHL) = O(N(HL+2)).$ For $T$ time steps, the computational  complexity is $O(N(HL+2)T).$

\subsection{Online rollout policy when multiple arms are played}
We now discuss rollout policy for  $M >1,$ more than one arm is played at each time-step. 

When a decision maker plays more than one arm  in each time-step in rollout policy, employing a base policy with future look-ahead is non-trivial. 

This is due to the large number of possible combinations of $M$ out of $N$ available arms, i.e., $\binom{N}{M}$. Since the rollout policy depends on future look-ahead actions, it can be computationally expensive to implement as each time-step because we need to choose from $\binom{N}{M}.$ 
We reduce these computations for base policy $\phi$ by employing a myopic rule in look-ahead approach, where we select $M$ arms with highest immediate rewards while computing value estimates of trajectories.

In this case, $\sum_{n=1}^{N}b_{h,l}^{n,\phi} = M,$  $M>1.$ The set of arms played at step $h$ in trajectory $l$ is  $\xi_{h,l} \subset \mathcal{N} = \{1,2, \cdots, N \},$ with $\vert \xi_{h,l} \vert = M.$  Here, $b_{h,l}^{n,\phi}  =1$ if $n\in\xi_{h,l}.$
The base policy $\phi$ uses myopic decision rule and the one-step policy improvement is given by 
\begin{eqnarray}
{j}^*(X) = \arg \max_{\xi \subset \mathcal{N}} \left[ \widetilde{r}(X, {\xi}) + \beta \widetilde{Q}_{H,L}^{\phi}(X, \xi) \right],
\label{eqn:policy-improv-2}
\end{eqnarray}
and $\widetilde{r}(X ,\xi) = \sum_{n=1}^{N} r^n(X^n, b^n).$ 
The computation of $\widetilde{Q}_{H,L}^{\phi}(X,\xi)$ is similar to the preceding discussion. 
At time $t$ with state $X = \{X^1,X^2, \cdots, X^N\},$  
rollout policy plays the subset of arms $j^*$  according to Eqn.~\eqref{eqn:policy-improv-2}.

\subsubsection{Computation complexity result for $M>1$ (multiple arms are played)}
Along  each trajectory, $M$ arms are played out of $N.$
The number of possible actions at each time-step in a  trajectory is   $\binom{N}{M}$ and  $\binom{N}{M} \approx O(N^M),$ which is a polynomial in $N$ for fixed $M.$ Thus, the complexity would be very high if all the possible actions are considered. 


The value estimates are computed only for some $|A|$ initial actions, $N\leq |A|< \binom{N}{M}.$  The computations required for value estimate $\widetilde{Q}_{H,L}^{\phi}(X, \xi)$ per iteration is $O(A H L).$ As there are $A$ number of  subsets considered for each step in action selection, the computations needed for policy improvement steps are at most $2|A|.$ So, the per-iteration computation complexity is $|A|HL + 2|A|.$ 
Hence, for $T$ time-steps the computational complexity would be $O(|A|(HL+2)T).$

\subsection{Additional Numerical Examples: Single-Armed Restless Bandit}
\label{sec:appendix-sab1}
Here we provide more numerical examples and verify the indexability using our Algorithm~\ref{algo:VI}. 
\subsubsection{Examples with restart}
We illustrate the structure of policy matrix $\Phi$ for restart model, where when action $1$ is taken, then $p^{1}_{s,1} = 1$ for all $s \in \mathcal{S}.$  Reward is decreasing in $s$ for passive action $(a =0)$ and reward for active action $(a =1)$ is $0$ for all states. We provide examples for $K = 10$ and $100$ states. The transition probabilities are as follows. $p^{0}(s, \min\{s+1,K \}) = 0.9$, $p^{0}(s, 1) = 0.1,$  $p^{1}(s,1) = 1$ for all $s \in \mathcal{S}.$

\begin{figure}[h]
	\begin{center} 
		\includegraphics[scale=0.28]{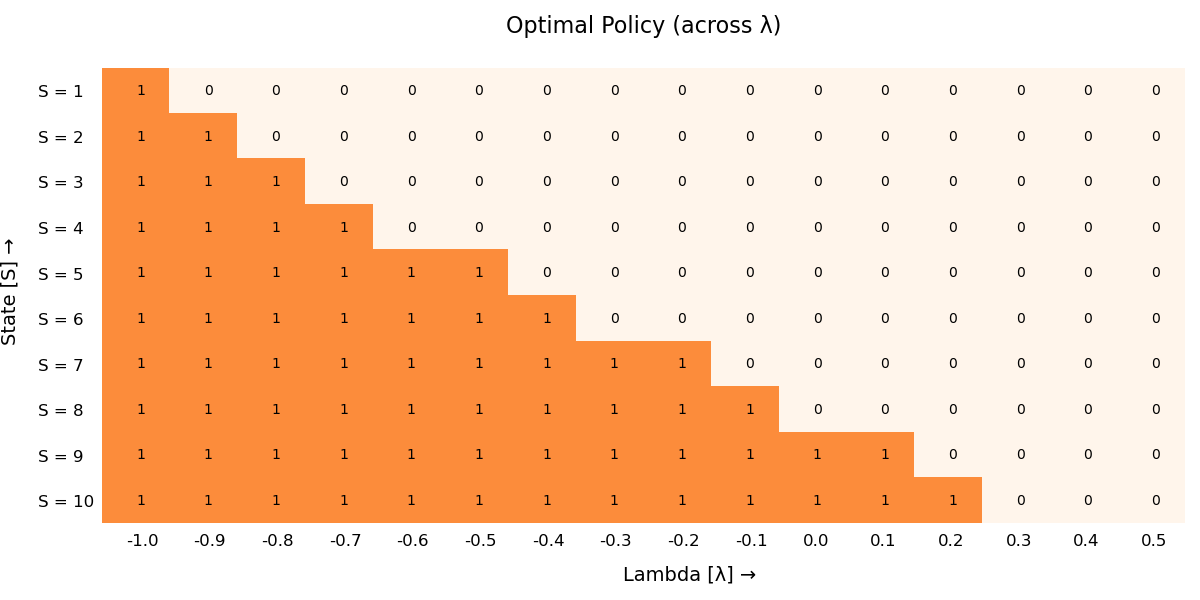}
	\end{center}
	\caption{Example of restart model: policy matrix $\Phi$ for different states as function of subsidy $\lambda.$ States $K=10.$}
	\label{fig:restart-model-10states}
\end{figure} 

For  $K=10,$ the rewards for action $a=0$ (passive action) with state $k,$ $r(k,0) = 0.95^k$ and the rewards for action $a=1$ (active action), $r(k,1) =0$ for all $k \in \mathcal{S}.$ The policy matrix $\Phi$ is illustrated in Fig.~\ref{fig:restart-model-10states}. We observe that the optimal policy is of a single threshold type and it is indexable.

\begin{figure}[h] 
	\begin{center} 
		\includegraphics[scale=0.3]{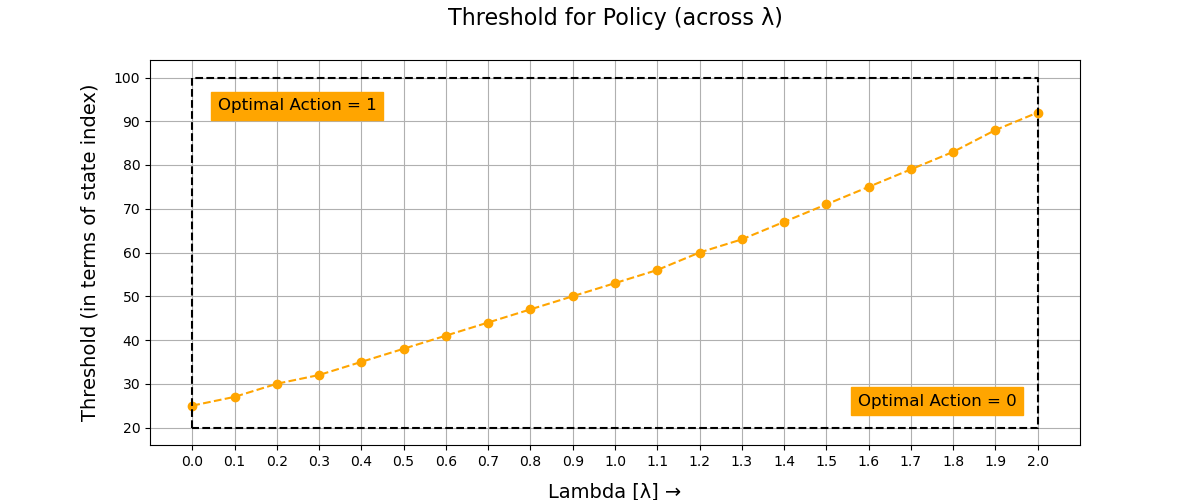}
	\end{center}
	\caption{Example of restart model: Threshold state $\hat{s}_{\lambda}$ as function of subsidy $\lambda.$ States $K=100.$}
	\label{fig:restart-model-100states}
\end{figure}

In Fig.~\ref{fig:restart-model-100states}, we use $K=100,$  and rewards for passive action as $r(k,0) = 0.99^k$ and the rewards for active action as $r(k,1) =0$ for all $k \in \mathcal{S}.$ It is difficult to draw policy matrix $\Phi$ for large states. 
Hence, we illustrate threshold state $\hat{s}_{\lambda}$ as function of $\lambda$. Here, threshold state  $\hat{s}_{\lambda}:= \inf  \{s: \pi_{\lambda}(s) = 0 \}.$ Notice that $\hat{s}_{\lambda}$ is increasing in $\lambda.$
This monotone behavior indicates  the bandit is indexable. 


\subsubsection{Example of one-step random walk}
We consider $K=5$ states. It is one-step random walk. The probability matrix is same for both the actions $a=1$ and $a=0$. Rewards for passive action $r(k,0) = 0$ and active action $r(k,1) = 0.9^k$ for all $k \in \mathcal{S}.$ This is also applicable in wireless communication systems. 

\begin{figure}[h] 
	\begin{center} 
		\includegraphics[scale=0.35]{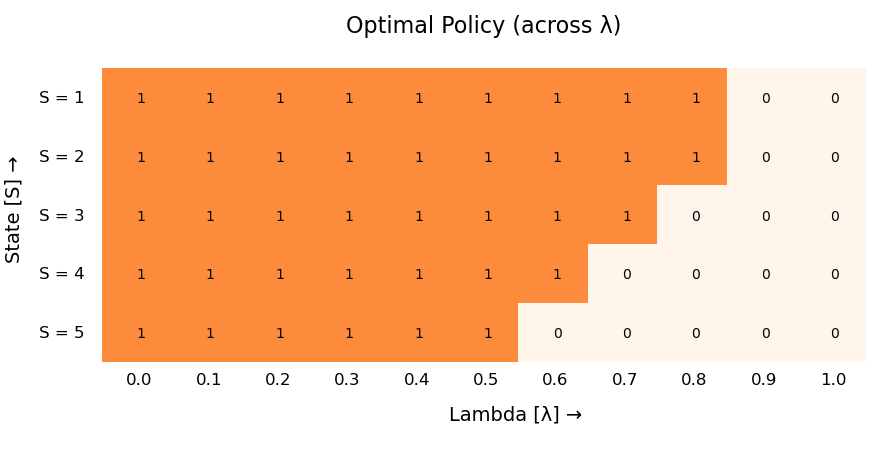}
	\end{center} 
	\caption{Example of one step random walk model: policy matrix $\Phi$ for different states as function of subsidy $\lambda.$ States $K=5.$}
	\label{fig:onestep-randomwalk-model-5state}
\end{figure}

\begin{eqnarray*}
	P_0 &=&  \begin{bmatrix}
		3/10 & 7/10 & 0 & 0 & 0 \\
		1/10 & 2/10 & 7/10 & 0 & 0 \\
		0 & 1/10 & 2/10 & 7/10 & 0 \\
		0 & 0 & 1/10 & 2/10 & 7/10 \\
		0 & 0 & 0 & 3/10 & 7/10
	\end{bmatrix}, \\
	P_1 &=& P_0.
\end{eqnarray*}

In Fig.~\ref{fig:onestep-randomwalk-model-5state}, we illustrate the policy matrix $\Phi$ and we observe that the policy $\pi_{s}(\lambda)$ is non-increasing in $s$ for fixed $\lambda.$ This is true for all $\lambda$. Thus, the policy has a single-threshold in $s$.  For fixed state $s$, $\pi_{\lambda}(s)$ is non-increasing in $\lambda$ and it is true for all $s \in \mathcal{S}$. Hence, it has a single threshold policy in $\lambda$ also. 

From Fig.~\ref{fig:onestep-randomwalk-model-5state}, it is clear that $B(\lambda = 0.5) = \emptyset,$ $B(\lambda = 0.6) =  \{5\},$ $B(\lambda = 0.7) =  \{4, 5\},$ $B(\lambda = 0.8) =  \{3,4,5\},$ $B(\lambda = 0.9) =  \{1,2,3,4,5\}$ and $B(\lambda_1) \subseteq B(\lambda_2)$ for $\lambda_2 > \lambda_1.$ Thus, it is indexable. 

\subsubsection{Example from \cite{Akbarzadeh2022} (Akbarzadeh $2022$) }

\begin{figure}[h] 
	\begin{center} 
		\includegraphics[scale=0.35]{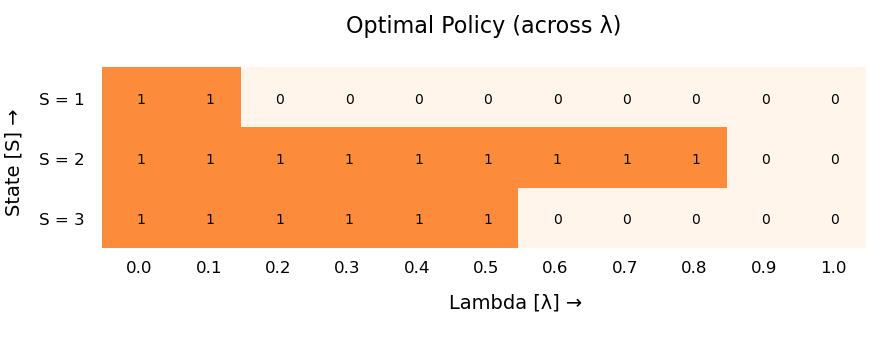}
	\end{center}
	\caption{Example of a model with no structure: policy matrix $\Phi$ for different states as function of subsidy $\lambda.$ States $K=3.$}
	\label{fig:Nostructure-model-3state-2action}
\end{figure} 

In this example, there is no structural assumption on transition probability and reward matrices. 
\begin{eqnarray*}
	P_0 &=&  \begin{bmatrix}
		0.3629 & 0.5026 & 0.1343 \\
		0.0823 & 0.7534 & 0.1643  \\
		0.2460 & 0.0294 & 0.7246 
	\end{bmatrix}, \\
	P_1 &=& 
	\begin{bmatrix}
		0.1719 & 0.1749 & 0.6532 \\
		0.0547 & 0.9317 & 0.0136 \\
		0.1547 & 0.6271 & 0.2182 
	\end{bmatrix}.
\end{eqnarray*}
Rewards in passive action $r(k,0) = 0$ for all $k \in \mathcal{S}$ and in active action, $r(1,1) = 0.44138$, $r(2,1) = 0.8033$, and $r(3,1)= 0.14257$. 

From policy matrix in Fig.~\ref{fig:Nostructure-model-3state-2action}, we can observe that the bandit is indexable, because $B(\lambda = 0.1) = \emptyset$, $B(\lambda = 0.2) = \{ 1\}$, $B(\lambda = 0.6) = \{ 1,3\},$ and $B(\lambda = 0.9) = \{ 1,2,3\}.$ $B(\lambda)$ is non-decreasing in $\lambda.$



\subsubsection{Indexable model from \cite{Nino-Mora07} (Nino-Mora $2007$)}

\begin{figure}[h] 
	\begin{center} 
		\includegraphics[scale=0.3]{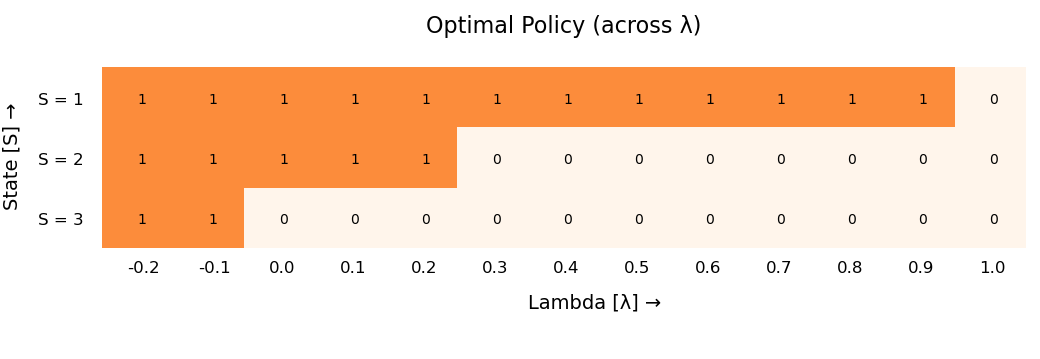}
	\end{center}
	\caption{Example of a model with no structure on reward and prob matrix from \cite{Nino-Mora07}: policy matrix $\Phi$ for different states as function of subsidy $\lambda.$ States $K=3.$}
	\label{fig:NinoMora-model-3state-2action}
\end{figure}

There is no structural assumption on transition probability matrices. 

\begin{eqnarray*}
	P_0 &=&  \begin{bmatrix}
		0.1810 & 0.4801 & 0.3389 \\
		0.2676 & 0.2646 & 0.4678  \\
		0.5304 & 0.2843 & 0.1853 
	\end{bmatrix}, \\
	P_1 &=& 
	\begin{bmatrix}
		0.2841 & 0.4827 & 0.2332 \\
		0.5131 & 0.0212 & 0.4657 \\
		0.4612 & 0.0081 & 0.5307 
	\end{bmatrix}, \\
	R &= &
	\begin{bmatrix}
		0 & 0.9016  \\
		0 & 0.10949  \\
		0 & 0.01055  
	\end{bmatrix}.
\end{eqnarray*}
We use the discount parameter as $\beta = 0.9.$ We observe from Fig.~\ref{fig:NinoMora-model-3state-2action}, that the optimal policy $\pi_{\lambda}(s)$ is non-increasing in $s$ for fixed $\lambda,$ even though there is no structural assumption on transition probability and reward matrices. Further, $\pi_{\lambda}(s)$ is non-increasing in $\lambda$ for fixed $s,$ This is true for all $s \in \mathcal{S}.$ Notice that $B(\lambda =-0.1) =\emptyset,$ $B(\lambda =0) =\{3\},$ $B(\lambda =0.3) =\{ 2,3 \},$ and $B(\lambda =1) =\{1,2,3 \}.$ Thus, the bandit is indexable.

\subsubsection{Non-indexable model from \cite{Nino-Mora07} (Nino-Mora $2007$)} 
\begin{figure}[h] 
	\begin{center} 
		\includegraphics[scale=0.2]{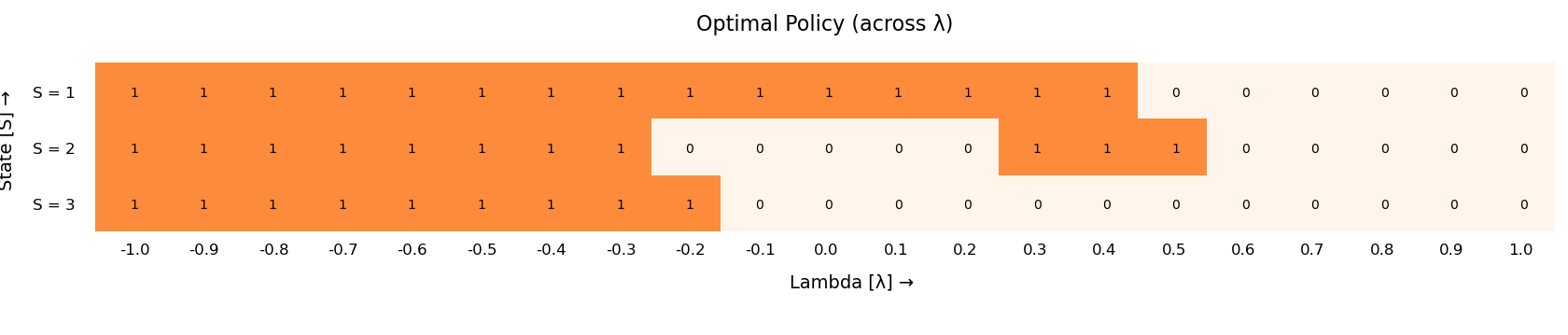}
	\end{center}
	\caption{Example of a model with no structure on reward and prob matrix from \cite{Nino-Mora07}: policy matrix $\Phi$ for different states as function of subsidy $\lambda.$ States $K=3.$ Non-indexable example.}
	\label{fig:NinoMora-model-3state-2action-Non-indexable}
\end{figure}

In this, we select an example of a bandit that  is non-indexable. We use the following  parameters. $\beta =0.9,$ 
\begin{eqnarray*}
	P_0 &=&  \begin{bmatrix}
		0.1902 & 0.4156 & 0.3942 \\
		0.5676 & 0.4191 & 0.0133  \\
		0.0191 & 0.1097 & 0.8712 
	\end{bmatrix}, \\
	P_1 &=& 
	\begin{bmatrix}
		0.7796 & 0.0903 & 0.1301 \\
		0.1903 & 0.1863 & 0.6234 \\
		0.2901 & 0.3901 & 0.3198 
	\end{bmatrix}, \\
	R &= &
	\begin{bmatrix}
		0.458 & 0.9631  \\
		0.5308 & 0.7963  \\
		0.6873 & 0.1057  
	\end{bmatrix}.
\end{eqnarray*}
We do not make any structural assumption on transition probability matrices but there is one on reward matrix. Reward is decreasing in $s$ for active action $(a = 1)$ and reward is increasing in $s$ for passive action $(a = 0)$.
In Fig.~\ref{fig:NinoMora-model-3state-2action-Non-indexable}, notice that policy $\pi_{\lambda}(s)$ is not monotone in $s$ for all values of $\lambda \in \Lambda.$ This implies that there exists $\lambda$ such that there are multiple thresholds, This can be observed for $\lambda = -0.2$, $\pi_{\lambda}(s) = [1,0,1]^T$  and for $\lambda = 0.5$, $\pi_{\lambda}(s) = [0,1,0]^T.$ Also, $\pi_{\lambda}(s)$
is not monotone in $\lambda$ for all values of $s \in \mathcal{S}.$ For instance, there is $s =2,$ for which $\pi_{\lambda}(s=2) = [1,0,0,0,0,0,1,1,1,0]$ for $\lambda =-0.3$ to $\lambda = 0.6$. 
$B(\lambda = -0.3) = \emptyset$, $B(\lambda = -0.2) = \{2\}$, $B(\lambda = -0.1) = \{2,3 \}$, $B(\lambda = 0.2) = \{2,3 \}$, $B(\lambda = 0.3) = \{3 \}$, $B(\lambda = 0.4) = \{3 \}$, $B(\lambda = 0.5) = \{1,3 \}$,
$B(\lambda = 0.6) = \{1,2,3 \}$.
Clearly, $B(\lambda)$ is not monotone in $\lambda,$ hence, the bandit is non-indexable.

\subsection{Non-indexable to indexable bandit by modification of reward matrix}
\label{sec:appendix-sab2}
In the following example, we use states $K=5$ and use the same transition probability matrices as in the non-indexable bandit. We modify here, the reward matrix and illustrate that the bandit becomes indexable.

\subsubsection{Indexable model with $5$ states and $\beta =0.9$} 

\begin{figure}[h] 
	\begin{center}
		\includegraphics[scale=0.18]{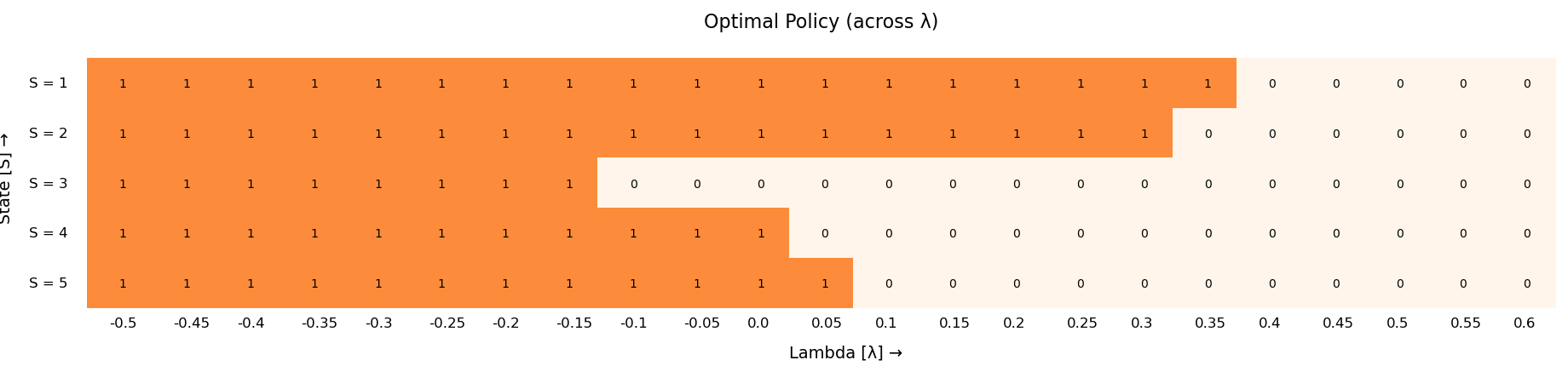}
	\end{center}
	\caption{Example of a model with no structure on reward and prob matrix : policy matrix $\Phi$ for different states as function of subsidy $\lambda.$ States $K=5.$ Bandit is indexable.}
	\label{fig:Our-model-5state-2action-indexable-modify-reward1}
\end{figure}

In this example, we slightly modify the reward matrix while keeping the same transition probability matrices as in the previous example, $K=5$. Then, we observe that such a bandit becomes indexable.
This may be due to the fact that the difference in reward from passive and action actions is decreased here as compared to the previous example, but the structure of reward matrix remains the same in both examples, i.e., reward is decreasing in $s$ for active action and reward is increasing in $s$ for passive action. 

\begin{eqnarray*}
	P_0 &=&  \begin{bmatrix}
		0.1502 & 0.0400 & 0.4156 & 0.0300 & 0.3642 \\
		0.4000 & 0.3500& 0.0800& 0.1200& 0.0500 \\
		0.5276 & 0.0400 &0.3991 & 0.0200 & 0.0133  \\
		0.0500 & 0.1000& 0.1500& 0.2000& 0.5000 \\
		0.0191 & 0.0100 & 0.0897 & 0.0300 & 0.8512 
	\end{bmatrix}, \\
	P_1 &=& 
	\begin{bmatrix}
		0.7196 & 0.0500 & 0.0903 & 0.0100 & 0.1301 \\
		0.5500 & 0.2000& 0.0500& 0.0800& 0.1200 \\ 
		0.1903 & 0.0100 & 0.1663 & 0.0100 & 0.6234 \\
		0.2000 & 0.0500 &0.1500 & 0.1000 & 0.5000 \\
		0.2501 & 0.0100& 0.3901 & 0.0300 & 0.3198 
	\end{bmatrix}, \\
	R &= &
	\begin{bmatrix}
		0.4580 & 0.9631  \\
		0.5100 & 0.8100 \\
		0.6508 & 0.7963  \\
		0.6710 & 0.6061 \\
		0.6873 & 0.5057  
	\end{bmatrix}.
\end{eqnarray*}
$\beta = 0.9$ In Fig.~\ref{fig:Our-model-5state-2action-indexable-modify-reward1}, we notice that $\pi_{\lambda}(s)$ is non-increasing (monotone) in $\lambda$ for each $s \in \mathcal{S}$ but $\pi_{\lambda}(s)$ is not monotone in $s$ for  $\lambda = -0.1, -0.05, 0.0, 0.05$. Further, we observe that $B(\lambda = -0.15) = \emptyset,$ $B(\lambda = -0.1) = \{ 3 \},$ $B(\lambda = 0.05) = \{ 3,4 \},$
$B(\lambda = 0.1) = \{ 3,4,5 \},$
$B(\lambda = 0.3) = \{ 3,4,5 \},$
$B(\lambda = 0.35) = \{ 2,3,4,5 \},$
$B(\lambda = 0.4) = \{ 1,2,3,4,5 \}.$
Hence, $B(\lambda)$ is non-decreasing in $\lambda$, and hence, it is indexable.

\subsubsection{Indexable model with $5$ states,  monotone reward  and $\beta =0.9$}
\begin{figure}[h] 
	\begin{center} 
		\includegraphics[scale=0.18]{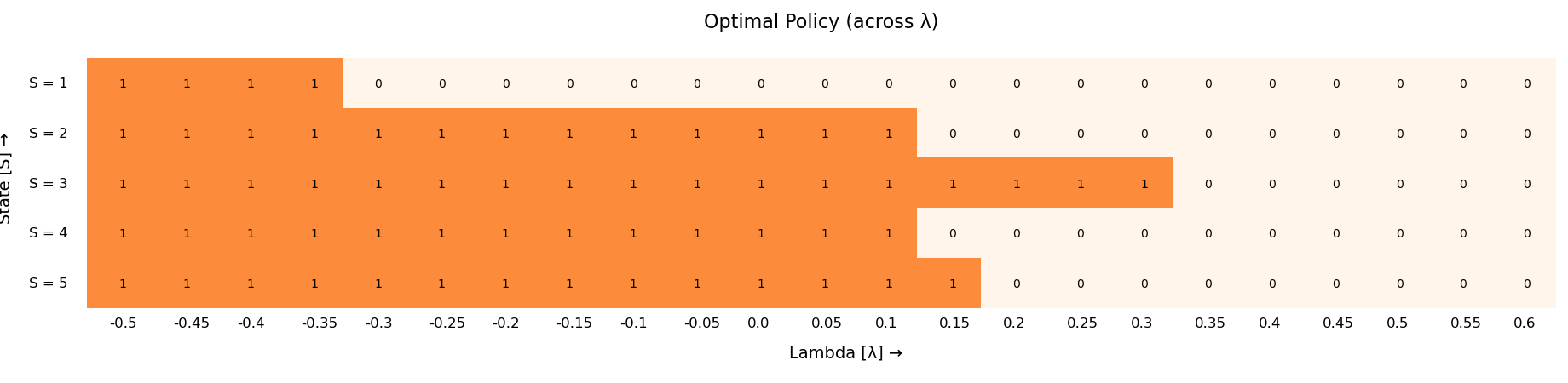}
	\end{center}
	\caption{Example of a model with no structure on reward and prob matrix : policy matrix $\Phi$ for different states as function of subsidy $\lambda.$ States $K=5.$ Bandit is indexable.}
	\label{fig:Our-model-5state-2action-indexable-modify-reward2}
\end{figure}

We consider following parameters.
\begin{eqnarray*}
	P_0 &=&  \begin{bmatrix}
		0.1502 & 0.0400 & 0.4156 & 0.0300 & 0.3642 \\
		0.4000 & 0.3500& 0.0800& 0.1200& 0.0500 \\
		0.5276 & 0.0400 &0.3991 & 0.0200 & 0.0133  \\
		0.0500 & 0.1000& 0.1500& 0.2000& 0.5000 \\
		0.0191 & 0.0100 & 0.0897 & 0.0300 & 0.8512 
	\end{bmatrix}, \\
	P_1 &=& 
	\begin{bmatrix}
		0.7196 & 0.0500 & 0.0903 & 0.0100 & 0.1301 \\
		0.5500 & 0.2000& 0.0500& 0.0800& 0.1200 \\ 
		0.1903 & 0.0100 & 0.1663 & 0.0100 & 0.6234 \\
		0.2000 & 0.0500 &0.1500 & 0.1000 & 0.5000 \\
		0.2501 & 0.0100& 0.3901 & 0.0300 & 0.3198 
	\end{bmatrix}, \\
	R &= &
	\begin{bmatrix}
		0.4580 & 0.5057  \\
		0.5100 & 0.6061 \\
		0.6508 & 0.7963  \\
		0.6710 & 0.8100 \\
		0.6873 & 0.9631  
	\end{bmatrix}.
\end{eqnarray*}
$\beta = 0.9.$ Reward is increasing in $s$ for both actions (passive $a=0$ and active $a=1$). But there is no structural assumption on transition probability matrices.  Using this example, we illustrate that the bandit is indexable.  From Fig,~\ref{fig:Our-model-5state-2action-indexable-modify-reward2}, we can observe that  $\pi_{\lambda}(s)$ is non-increasing in $\lambda$ for every $s \in \mathcal{S}$ but $\pi_{\lambda}(s)$ is not monotone in $S$ for $\lambda = 0.15, 0.2, 0.25, 0.3$. There is no single threshold policy in $s$ for these values of $\lambda$, but there is a single threshold policy in $\lambda$ for each $s \in \mathcal{S}$. We  have $B(\lambda = -0.35) = \emptyset$, $B(\lambda = -0.3) = \{1\}$, $B(\lambda = 0.15) = \{1,2,4\}$, $B(\lambda = 0.2) = \{1,2,4,5\}$ and $B(\lambda = 0.35) = \{1,2,3,4,5\}$. Set $B(\lambda)$ is non-decreasing in $\lambda$ and hence, the bandit is indexable.  

\subsubsection{Indexable model with $5$ states and $\beta = 0.99$}

\begin{figure}[h] 
	\begin{center} 
		\includegraphics[scale=0.18]{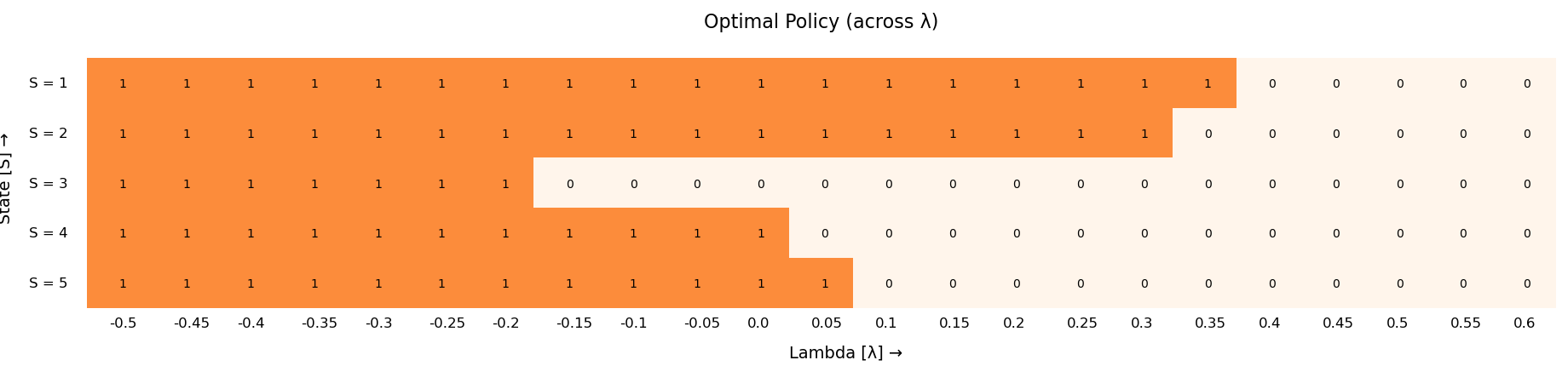}
	\end{center} 
	\caption{Example of a model with no structure on reward and prob matrix : policy matrix $\Phi$ for different states as function of subsidy $\lambda.$ States $K=5$ and $\beta = 0.99,$ Bandit is indexable.}
	\label{fig:Our-model-5state-2action-indexable-modify-reward1-beta}
\end{figure}

Consider 

\begin{eqnarray*}
	P_0 &=&  \begin{bmatrix}
		0.1502 & 0.0400 & 0.4156 & 0.0300 & 0.3642 \\
		0.4000 & 0.3500& 0.0800& 0.1200& 0.0500 \\
		0.5276 & 0.0400 &0.3991 & 0.0200 & 0.0133  \\
		0.0500 & 0.1000& 0.1500& 0.2000& 0.5000 \\
		0.0191 & 0.0100 & 0.0897 & 0.0300 & 0.8512 
	\end{bmatrix}, \\
	P_1 &=& 
	\begin{bmatrix}
		0.7196 & 0.0500 & 0.0903 & 0.0100 & 0.1301 \\
		0.5500 & 0.2000& 0.0500& 0.0800& 0.1200 \\ 
		0.1903 & 0.0100 & 0.1663 & 0.0100 & 0.6234 \\
		0.2000 & 0.0500 &0.1500 & 0.1000 & 0.5000 \\
		0.2501 & 0.0100& 0.3901 & 0.0300 & 0.3198 
	\end{bmatrix}, \\
	R &= &
	\begin{bmatrix}
		0.4580 & 0.9631  \\
		0.5100 & 0.8100 \\
		0.6508 & 0.7963  \\
		0.6710 & 0.6061 \\
		0.6873 & 0.5057  
	\end{bmatrix}.
\end{eqnarray*}
$\beta = 0.99.$ We now observe that the reward is monotone in $s$ for both actions. From Fig,~\ref{fig:Our-model-5state-2action-indexable-modify-reward1-beta}, we have $B(\lambda = -0.2) = \emptyset,$ $B(\lambda = -0.15) = \{3\},$ $B(\lambda = 0.05) = \{3,4\},$ $B(\lambda = 0.1) = \{3,4,5\},$
$B(\lambda = 0.35) = \{2,3,4,5\},$
and  $B(\lambda = 0.4) = \{1,2,3,4,5\},$
$B(\lambda)$ is non-decreasing in $\lambda.$ Hence, the bandit is indexable. 

\subsection{Discussion on non-indexability and indexability of SAB}

From the preceding  examples (non-indexable and indexable) in Section~\ref{sec:numerical-indexability-sab},  and Appendix ~\ref{sec:appendix-sab2},   we infer that in order for a bandit to be non-indexable, there is necessity of reward for both actions to be in the reverse order along the actions and there should be sufficient difference in rewards for each state. 
When reward is monotone (non-decreasing) in state for both actions, then the bandit is indexable, and this is due to, as we increase subsidy $\lambda$ there is no possibility of more than a single threshold in $\lambda$ for each state $s \in \mathcal{S}$. In order to have more than one threshold in $\lambda$, first we observe that $\lambda$ is fixed reward obtained for passive action and independent of state; second as $\lambda$ increases, passive action becomes optimal as not playing is optimal choice;  third as $\lambda$ increases more then active action to be optimal again, it requires cumulative reward from active action to be higher than that of passive action. 

It is possible due to the fact that for passive action, reward is increasing and for active action, it is decreasing. The difference $r(s,1)-r(s,0)$ is going from positive to negative value and increasing $\lambda$ provides balancing for some states by dynamic program equation. Fourth, we see that as $\lambda$ increases even further, the optimal action has to be passive and remains passive for remaining values of $\lambda$.   

\subsubsection{Comments on Indexability}

From numerical examples of  restless single-armed bandits, we observed  that non-indexability  occurs under very restrictive setting on transition probability and reward matrices. Most applications in wireless communication, machine maintenance, recommendation systems models, there is assumed to be some structure on reward  and transition probability matrices. Hence, many applications  of finite state observable restless bandit models are indexable. 

\subsection{Additional Numerical Examples: RMAB with Performance of different policies}
Here, we provide more numerical examples, comparing the performance of different policies on restless multi-armed bandits. In the following, we consider both, identical as well as non-identical restless bandits. Also, we use varying scenarios such as; monotone and indexable bandits; non-monotone and indexable bandits; and non-indexable bandits. In the scenario of non-indexable bandits, we  compare the performance of rollout and myopic policies.
We illustrate simulations for number of arms $N = 3, 5, 10$, and discount parameter $\beta = 0.99$. In the rollout policy, we used number of trajectories $L = 30$, and horizon length $H = 3, 4, 5$.

\subsubsection{Example of identical, indexable, and  monotone bandits}

\begin{figure}[h]
	\begin{center}
		\includegraphics[scale=0.35]{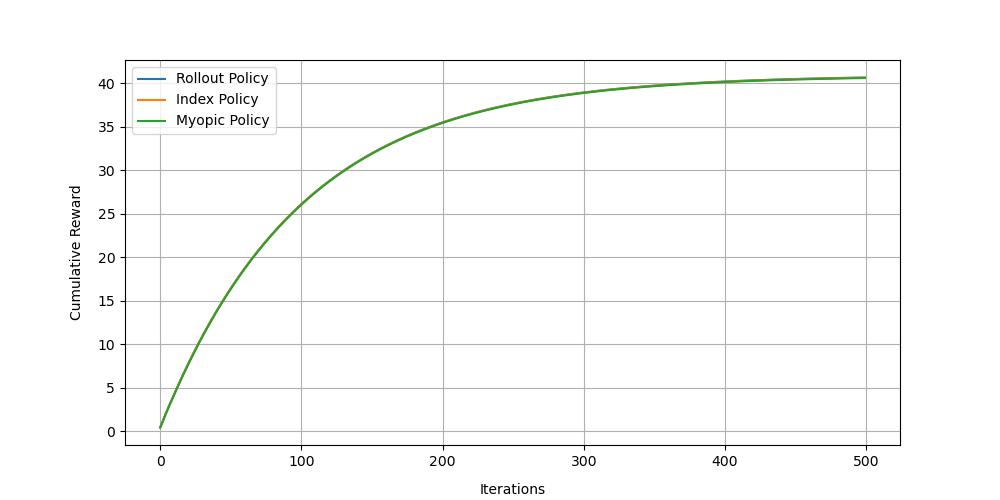}
	\end{center} 
	\caption{$5$ armed restless bandits: Identical, indexable and monotone.}
	\label{fig:RMAB-4}
\end{figure}

\begin{figure}[h]
	\begin{center} 
		\includegraphics[scale=0.35]{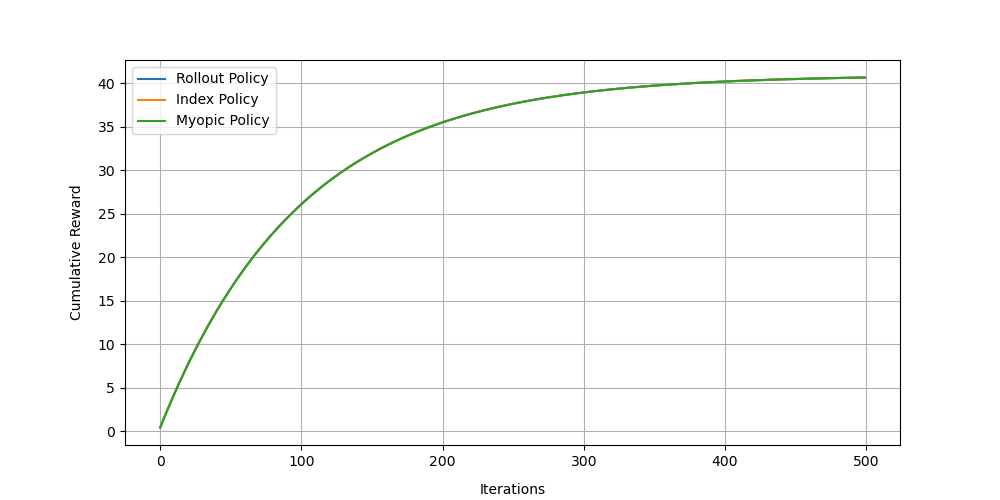}
	\end{center} 
	\caption{$10$ armed restless bandits: Identical, indexable and monotone.}
	\label{fig:RMAB-5}
\end{figure}

In Fig.~\ref{fig:RMAB-4} and Fig.~\ref{fig:RMAB-5}, we compare performance of myopic, rollout and index policy for $N = 5, 10$ respectively. All the bandits are identical, indexable, and index is monotone in state. We use number of trajectories $L = 30$, and horizon length $H = 4$. Notice that all policies have identical performance and it is due to  identical and monotone reward structure of bandits and index is monotone in state for the bandit.   

\subsubsection{Example of identical, indexable, and non-monotone bandits}

\begin{figure} 
	\begin{center} 
		\includegraphics[scale=0.35]{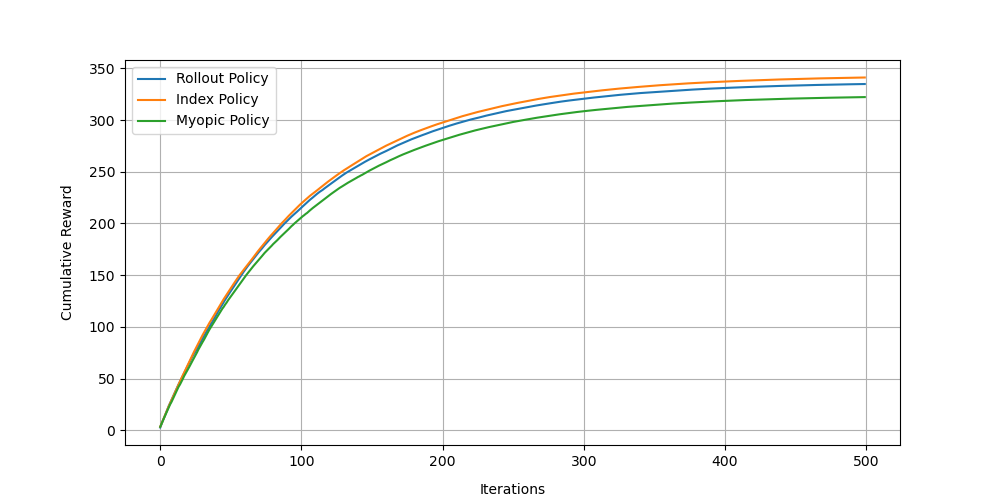}
	\end{center} 
	\caption{$5$ armed restless bandits: Identical, indexable and non monotone}
	\label{fig:RMAB-6}
\end{figure}

\begin{figure}
	\begin{center} 
		\includegraphics[scale=0.35]{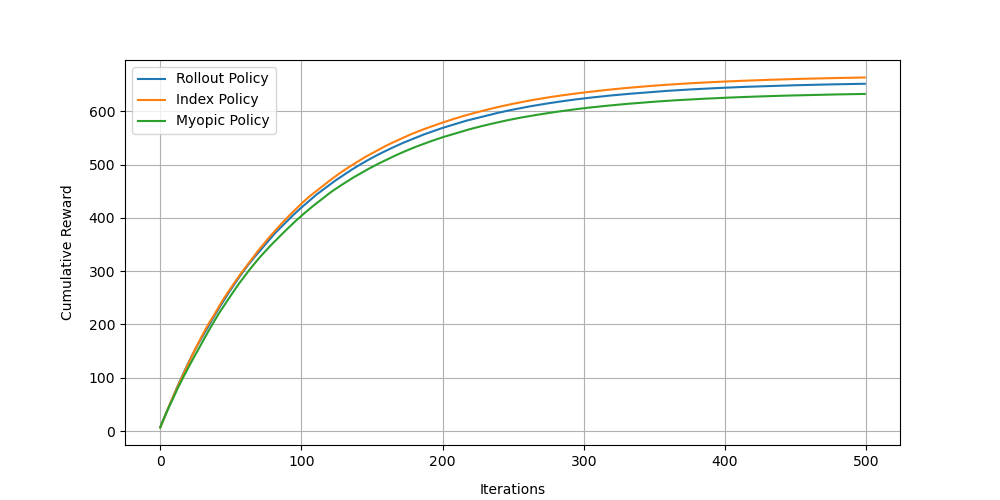}
	\end{center} 
	\caption{$10$ armed restless bandits: Identical, indexable and non monotone}
	\label{fig:RMAB-7}
\end{figure}
In Fig.~\ref{fig:RMAB-6} and Fig.~\ref{fig:RMAB-7}, we compare performance of myopic, rollout and index policy for $N = 5, 10$ respectively. All the bandits are identical, indexable, and non-monotone, i.e., index is not monotone in state. We use number of trajectories $L = 30$, and horizon length $H = 4$. We observe that the rollout policy and the index policy performs better than myopic policy. 

Examples of such bandit are given in  Appendix~\ref{sec:appendix-sab2}. Even though $B(\lambda)$ is non-decreasing in $\lambda,$ the policy $\pi_{\lambda}(s)$ is not monotone in $s$ for all values of $\lambda.$ The  examples from SAB suggests that there may not be  any structure on transition probability  or reward matrices. In such examples, myopic policy does not perform as good as rollout policy and index policy. Here, rollout policy uses simulation based look-ahead approach and index policy uses dynamic program for index computation, these policies take into account future possible state informations.

\begin{figure}
	\begin{center} 
		\includegraphics[scale=0.35]{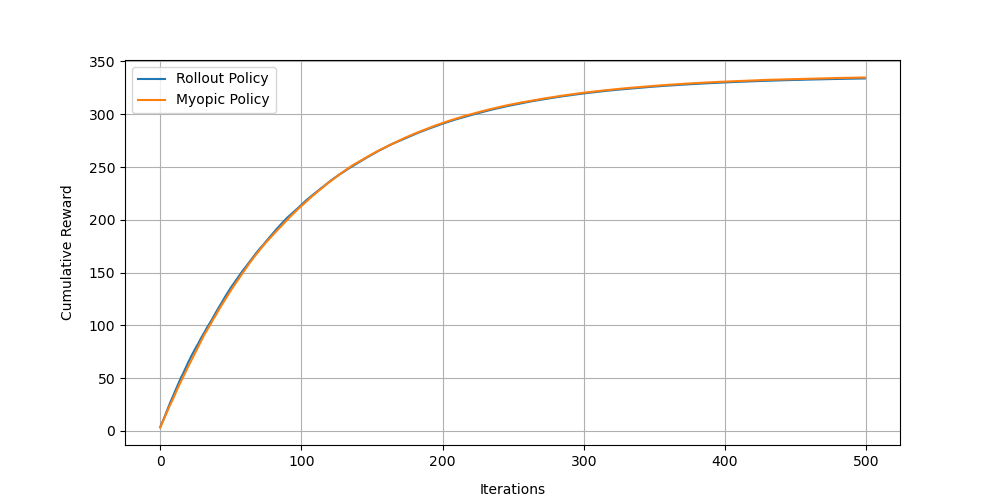}
	\end{center} 
	\caption{$5$ armed restless bandits: Identical  and non-indexable}
	\label{fig:RMAB-8}
\end{figure}

\begin{figure}
	\begin{center} 
		\includegraphics[scale=0.35]{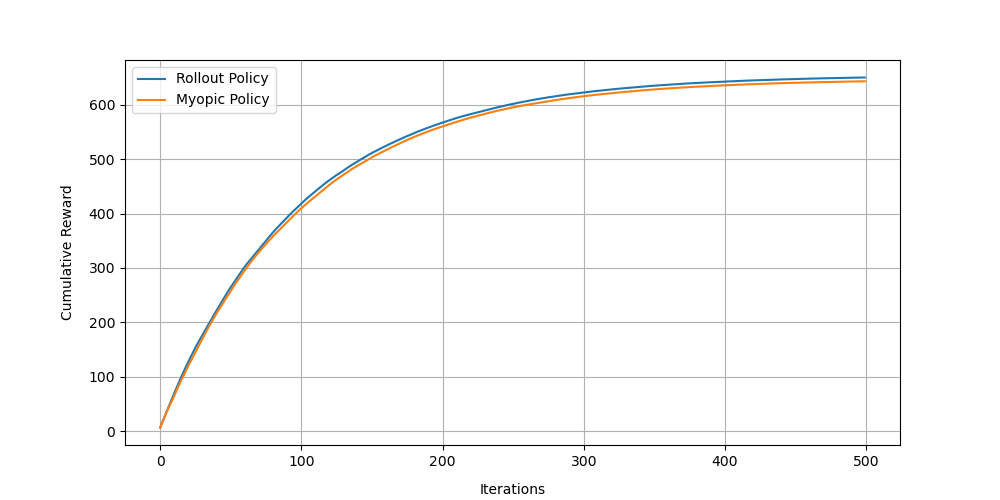}
	\end{center} 
	\caption{$10$ armed restless bandits: Identical  and non-indexable.}
	\label{fig:RMAB-9}
\end{figure}

\begin{figure}
	\begin{center} 
		\includegraphics[scale=0.35]{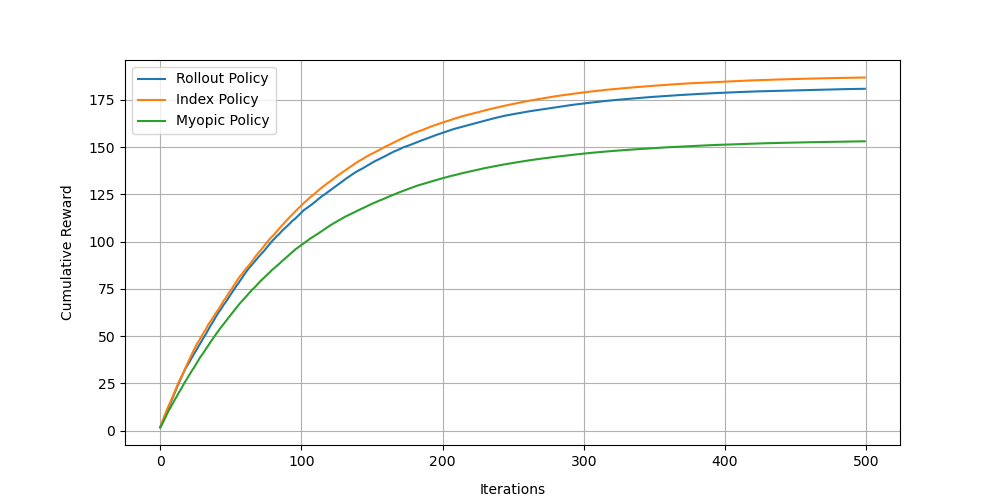}
	\end{center} 
	\caption{$3$ armed restless bandits: Non-Identical and indexable, $H =3.$}
	\label{fig:RMAB-10}
\end{figure}

\begin{figure}
	\begin{center} 
		\includegraphics[scale=0.37]{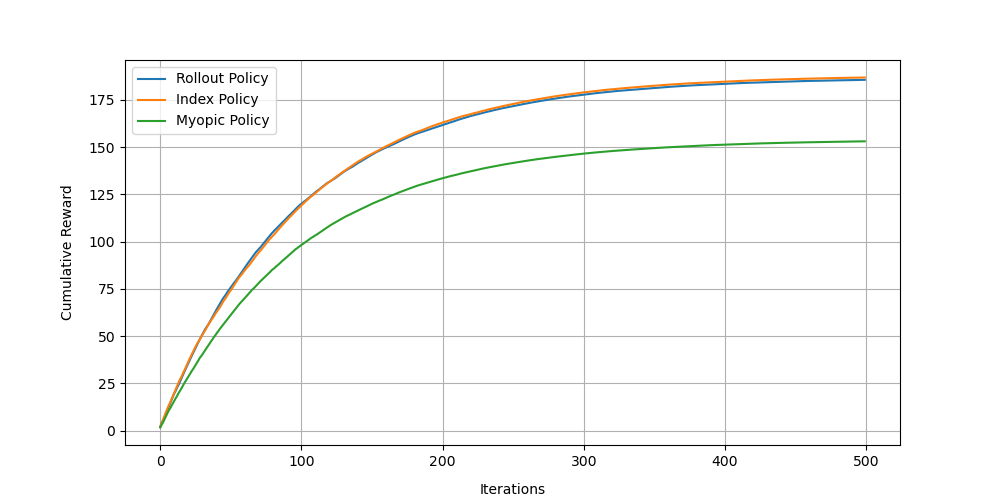}
	\end{center} 
	\caption{$3$ armed restless bandits: Non-identical  and indexable, $H =5.$}
	\label{fig:RMAB-11}
\end{figure}


\subsubsection{Example of  identical and non-indexable bandits}
In Fig.~\ref{fig:RMAB-8} and Fig.~\ref{fig:RMAB-9}, we compare performance of myopic and rollout policy for $N = 5, 10$ respectively, where we assume that all bandits are non-indexable and all bandits are identical. In the rollout policy, the number of trajectories are $L = 30$, and horizon length is $H = 4$. We observe that both policies have identical performance.  


\subsubsection{Effect of  horizon length $H$ in rollout policy}
In Fig.~\ref{fig:RMAB-10} and Fig.~\ref{fig:RMAB-11}, we compare performance of myopic, rollout and index policy for $N = 3$. All the bandits are non-identical and indexable, some are monotone, and some are non-monotone. We use number of trajectories $L = 30$, and horizon length $H = 3, 5$ respectively. We observe that the rollout policy and the index policy performs better than myopic policy. It is noticed that as $H$  increases from $3$ to $5,$ the difference in rollout policy and index policy decreases.

\end{document}